\documentclass[conference, twocolumn]{IEEEtran}
\usepackage{graphicx}
\usepackage{array}

\usepackage[utf8]{inputenc}
\usepackage[T1]{fontenc}

\usepackage{listings}
\usepackage{subcaption}
\usepackage[table,xcdraw]{xcolor}
\usepackage{amsmath}
\usepackage{bbm}
\usepackage{pifont}
\usepackage{amssymb}
\definecolor{mydarkblue}{rgb}{0,0.08,0.65}
\usepackage[colorlinks=true,
    linkcolor=mydarkblue,
    citecolor=mydarkblue,
    filecolor=mydarkblue,
    urlcolor=mydarkblue]{hyperref}
\usepackage{cleveref}
\usepackage{soul}
\usepackage[shortlabels]{enumitem}
\usepackage{url}
\usepackage{color}
\usepackage{tikz}
\usepackage{balance}
\usepackage{multirow}
\usepackage{comment}
\usepackage{wrapfig,booktabs}

\usepackage[symbol]{footmisc}
\usepackage{tablefootnote}
\usepackage{tabularx}
\usepackage{placeins}
\usepackage{algorithm}
\usepackage{algpseudocode}
\usepackage{ragged2e}
\usepackage{cuted}
\usepackage{float}
\usepackage{caption}

\usepackage{multicol}

\usepackage{dblfloatfix}
\usepackage{natbib}

\renewcommand{\arraystretch}{1.2}

\definecolor{codegreen}{rgb}{0,0.6,0}
\definecolor{codegray}{rgb}{0.5,0.5,0.5}
\definecolor{codepurple}{rgb}{0.58,0,0.82}
\definecolor{backcolour}{rgb}{0.95,0.95,0.92}

\makeatletter
\def\blfootnote{\xdef\@thefnmark{}\@footnotetext}
\makeatother

\lstdefinestyle{mystyle}{
  backgroundcolor=\color{backcolour},   commentstyle=\color{codegreen},
  keywordstyle=\color{magenta},
  numberstyle=\tiny\color{codegray},
  stringstyle=\color{codepurple},
  basicstyle=\ttfamily\footnotesize,
  breakatwhitespace=false,
  breaklines=true,
  captionpos=b,
  keepspaces=true,
  numbers=left,
  numbersep=5pt,
  showspaces=false,
  showstringspaces=false,
  showtabs=false,
  tabsize=2,
}

\AtBeginDocument{%
  \providecommand\BibTeX{{%
    \normalfont B\kern-0.5em{\scshape i\kern-0.25em b}\kern-0.8em\TeX}}}

\IEEEoverridecommandlockouts
\begin{document}

\title{ZAYA1-8B Technical Report}

\newcommand{\corr}{\textsuperscript{*}}

\author{
Robert Washbourne\corr, Rishi Iyer, Tomas Figliolia, Henry Zheng, Ryan Lorig-Roach,\\ Sungyeon Yang, Pritish Yuvraj, Quentin Anthony, Yury Tokpanov, Xiao Yang, Ganesh Nanduru,\\ Stephen Ebert, Praneeth Medepalli, Skyler Szot, Srivatsan Rajagopal, Alex Ong, Bhavana Mehta, \\ Beren Millidge\corr
\\[0.5em]
\textbf{Zyphra}
\\
San Francisco, CA \\
\IEEEauthorblockA{\textsuperscript{*}Corresponding authors: \texttt{rob@zyphra.com}, \texttt{beren@zyphra.com}}
}

\maketitle

\setcounter{page}{1}

\begin{abstract}\normalfont\mdseries
We present ZAYA1-8B, a reasoning-focused mixture-of-experts (MoE) model with 700M active and 8B total parameters, built on Zyphra's MoE++ architecture. ZAYA1-8B's core pretraining, midtraining, and supervised fine-tuning (SFT) were performed on a full-stack AMD compute, networking, and software platform. With under 1B active parameters, ZAYA1-8B
matches or exceeds DeepSeek-R1-0528 on several challenging
mathematics and coding benchmarks, and remains competitive with substantially larger open-weight reasoning models. ZAYA1-8B was trained from scratch for reasoning, with reasoning data included from pretraining onward using an answer-preserving trimming scheme. Post-training uses a four-stage RL cascade: reasoning warmup on math and puzzles; a 400-task RLVE-Gym curriculum; math and code RL with test-time compute traces and synthetic code environments built from competitive-programming references; and behavioral RL for chat and instruction following. We also introduce Markovian RSA, a test-time compute method that recursively aggregates parallel reasoning traces while carrying forward only bounded-length reasoning tails between rounds. In TTC evaluation, Markovian RSA raises ZAYA1-8B to 91.9\% on AIME'25 and 89.6\% on HMMT'25 while carrying forward only a 4K-token tail, narrowing the gap to much larger reasoning models including Gemini-2.5 Pro, DeepSeek-V3.2, and GPT-5-High.
\end{abstract}

\section{Introduction}
\label{sec:intro}
    

\begin{figure*}
    \centering
    \includegraphics[width=1\linewidth]{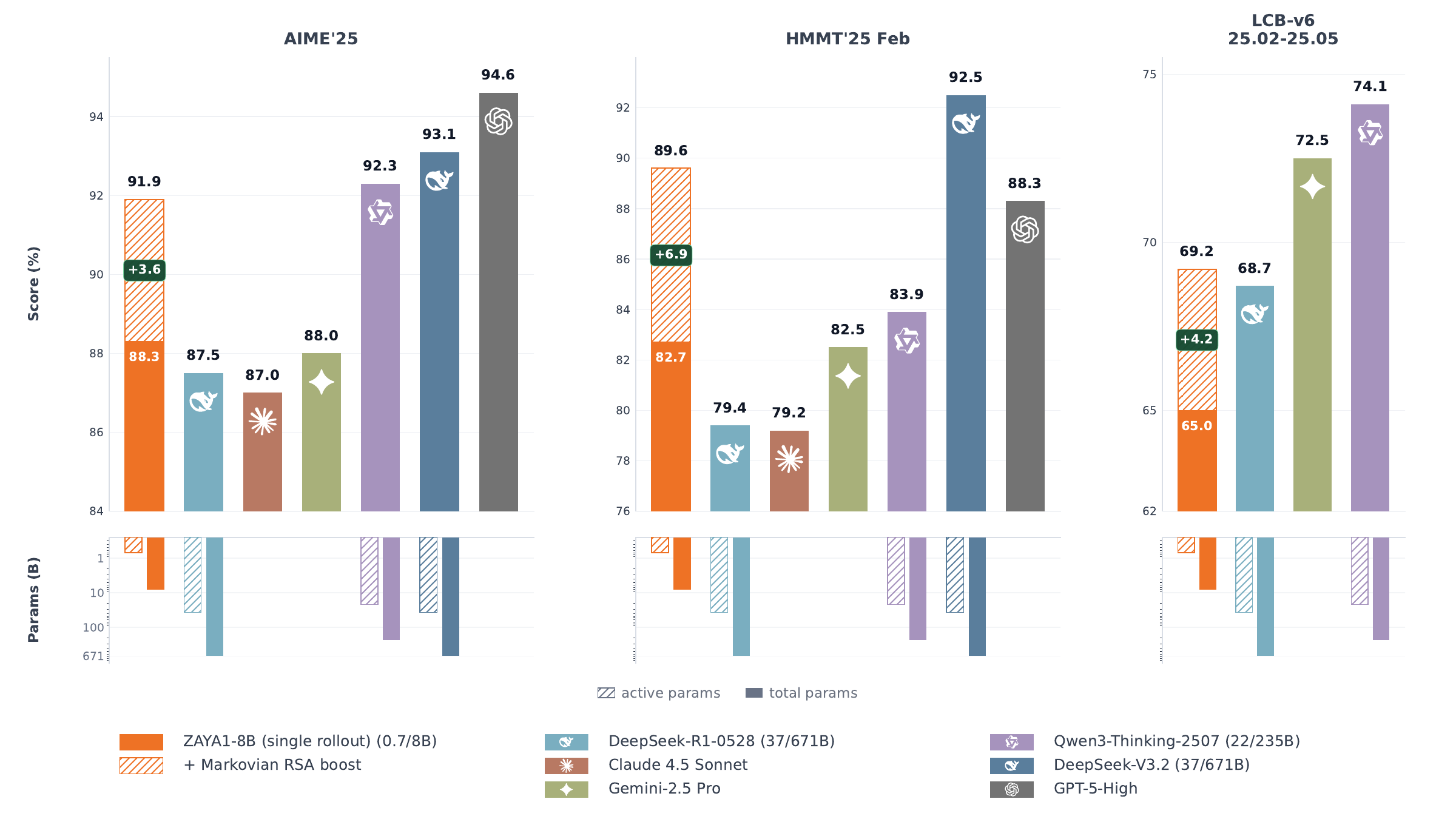}
    \caption{ZAYA1-8B with Markovian RSA test-time compute vs. substantially larger reasoning models on AIME'25, HMMT'25, and LCB-v6. Hatched bars show the boost from Markovian RSA over single-rollout ZAYA1-8B. With 0.7B active parameters and the 40K/4K Markovian RSA configuration (Section~\ref{sec:ttc-inference}), ZAYA1-8B reaches 91.9\% on AIME'25 and 89.6\% on HMMT'25, narrowing the gap to larger proprietary and open-weight reasoning models. ZAYA1-8B numbers (single-rollout and TTC) are evaluated in the Zyphra harness on the pre-behavioral checkpoint after math+code+TTC RL and before the final lightweight behavioral-RL polishing stage; comparator numbers are taken from official release materials (see Table~\ref{tab:results-ttc} for sources). The final behavioral stage targets chat style, instruction following, and preference behavior rather than math/code/TTC capability.}
    \label{markovian_rsa_bar}
\end{figure*}

\begin{figure*}[!t]
    \centering
    \includegraphics[width=\textwidth]{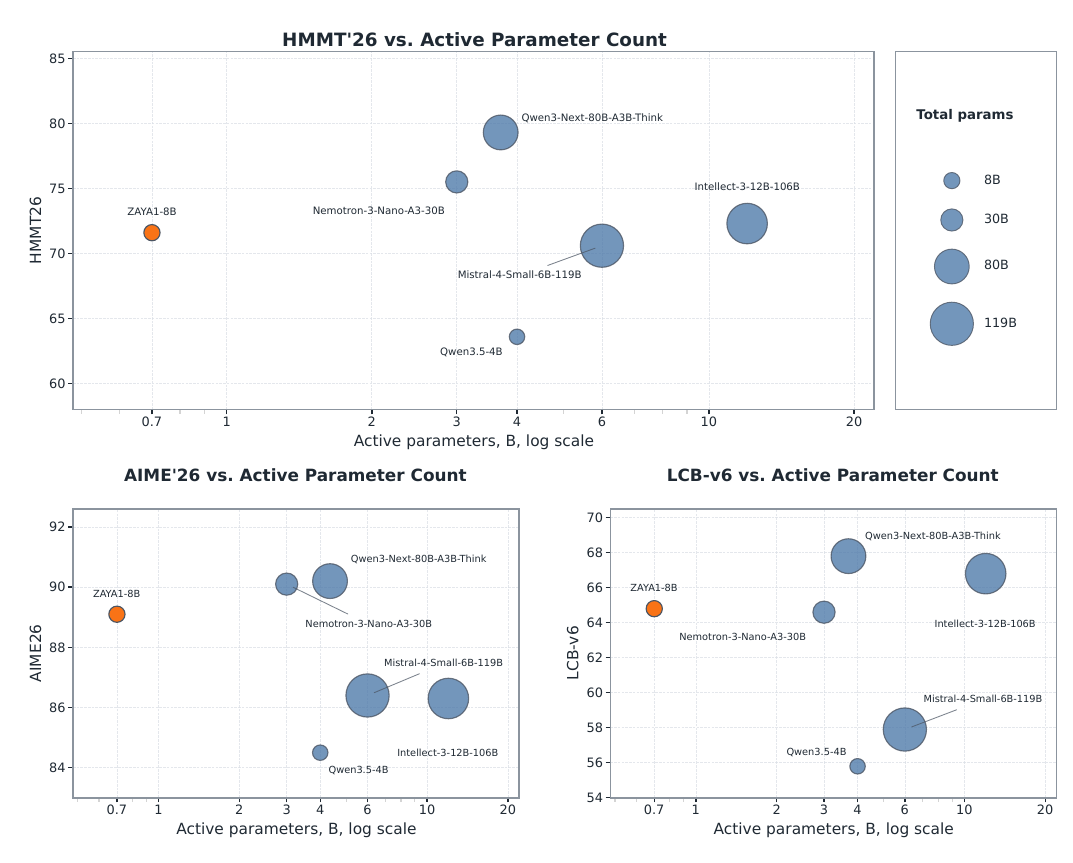}
    \caption{Active-parameter scaling across HMMT'26, AIME'26, and LiveCodeBench-v6. ZAYA1-8B is shown at 0.7B active parameters and compared against larger open-weight and frontier models where available. Bubble area denotes total parameter count where available.}
    \label{fig:active-params-combined}
\end{figure*}

In this paper, we introduce ZAYA1-8B, a 700M-active, 8B-total parameter mixture-of-experts (MoE) model. With under 1B active parameters, ZAYA1-8B matches or exceeds DeepSeek-R1-0528 on several challenging mathematics and coding benchmarks, while remaining competitive with substantially larger open-weight reasoning models including OLMo-3.1-32B-Think, Nemotron-3-Nano-30B-A3B, Mistral-Small-4-119B-2603, and Intellect-3-12A-106B \citep{nvidia_nemotron_nano_v3_2025,team2025olmo,intellect3,mistral_small_4_2026}.

Moreover, using our test-time compute scheme, Markovian RSA, ZAYA1-8B narrows the gap on AIME'25 and HMMT'25 to substantially larger reasoning models including Gemini-2.5 Pro, DeepSeek-V3.2, Qwen3-235B-A22B-Thinking-2507, and GPT-5-High \citep{comanici2025gemini,deepseekai2025deepseekv32,qwen3technicalreport,openai_2025_gpt5_system_card}. These results suggest that competitive mathematical reasoning can be reached with under 1B active parameters when model architecture, reasoning-heavy training, verifiable RL, and test-time aggregation are co-designed.

The system combines five design choices that we found important in practice:

\textbf{Architecture:} ZAYA1-8B builds on Zyphra's MoE++ architecture \citep{prior-zyphra-work}, with three main changes relative to standard transformer MoE designs. First, ZAYA1-8B uses Compressed Convolutional Attention (CCA) \citep{cca}, a FLOP- and memory-efficient attention variant that performs sequence mixing in a compressed latent space. Prior work showed that CCA performs well on perplexity and standard language modeling at small scale; ZAYA1-8B evaluates its behavior at larger scale and on more challenging reasoning and long-context tasks. Second, ZAYA1-8B uses the ZAYA1 router, which replaces the standard linear MoE router with a multi-layer MLP-based design, substantially increasing its expressiveness. In our experiments we find that increasing router capacity and expressiveness is a strong use of marginal parameters. A small number of router parameters controls a much larger number of expert parameters, and better routing decisions significantly reduce balancing instability and improve model quality. Third, ZAYA1-8B applies learned residual scaling to both the residual stream and the layer input at each block, which controls residual-norm growth through depth at negligible parameter and FLOP cost.

\textbf{Reasoning-aware training across stages:} 
We designed ZAYA1-8B from scratch for reasoning. Motivated by evidence that including reasoning data during pretraining can produce gains that post-training alone does not recover \citep{nvidia-synergy}, we include long chain-of-thought (CoT) data in all pretraining phases and during midtraining. To train on reasoning traces that exceed the pretraining context length, we introduce a novel answer-preserving trimming methodology, which truncates the tail of the reasoning trace while preserving the final answer, or drops the example if the answer alone does not fit. Unlike prior length-control methods that operate during inference or RL rollout generation \citep{scalerl, yang2025qwen3}, AP-trimming is applied during training-data construction.

\textbf{Cascaded reinforcement learning pipeline:} Post-training for ZAYA1-8B uses a four-stage RL cascade: reasoning warmup, a 400-task adaptive difficulty curriculum over the RLVE-Gym environment suite \citep{rlve-gym}, math and code RL with test-time compute traces, and a final behavioral RL stage. The cascade uses asynchronous PipelineRL \citep{pipelinerl, scalerl} with DPPO Binary-TV trust-region masking \citep{dppo}, Dr-GRPO sequence-level loss aggregation \citep{dr-grpo}, MaxRL advantage estimation~\citep{tajwar2026maxrl}, and no KL regularization in the reward. Stable training required substantial precision, verifier, and data curation work, which we document throughout the report.

\textbf{Test-time compute methods:} We introduce Markovian RSA, a novel test-time compute method that combines the recursive candidate-aggregation structure of RSA~\citep{rsa} with the bounded-workspace principle of Markovian Thinking~\citep{markovian-thinker}. Markovian RSA turns long reasoning into staged batched inference: each stage generates $N$ candidates in parallel, each candidate has bounded decode length $\beta$, and aggregation prefill depends only on $C$ carried-forward tails of length $\tau$, not on the full reasoning history. Crucially, we also integrate Markovian RSA into training: SFT data is constructed by reshuffling expert-model rollouts into aggregation examples, and RL stages train both expert-model and policy-self-aggregation variants. The resulting model is trained for the Markovian RSA workflow at inference and we achieve substantial performance uplift by doing so.

\textbf{AMD training stack:} Building on our prior work with AMD MI300X GPUs and AMD Pensando Pollara 400 networking for large-scale pretraining \citep{prior-zyphra-work}, ZAYA1-8B was pretrained, midtrained, and supervised fine-tuned on this GPU/networking stack. This provides evidence that the stack can support sustained pretraining, long-context midtraining, and supervised fine-tuning for an 8B-total-parameter MoE reasoning model. We validate this stack at the ZAYA1-8B scale; validation for substantially larger models and broader parallelism regimes remains future work.

The remainder of this report is organized as follows:
Section~\ref{sec:model} describes the ZAYA1-8B architecture. Section~\ref{sec:pretraining} describes pretraining, midtraining, and answer-preserving trimming. Section~\ref{sec:posttraining} describes the SFT stage and RL cascade, including infrastructure, precision, optimizer, and stability-monitoring choices. Section~\ref{sec:results} reports benchmark results and comparisons. 
Section~\ref{sec:ttc} describes our test-time compute approach. Section~\ref{sec:discussion} concludes with observations from training and open questions.

\section{Model}
\label{sec:model}

\begin{table*}[h]
\centering
\small
\begin{tabular}{ll}
\toprule
Property & ZAYA1-8B configuration \\
\midrule
Architecture family & Decoder-only MoE Transformer, Zyphra MoE++ \\
Active parameters & 0.76B \\
Total parameters & 8.4B \\
Transformer layers & 40 \\
Hidden dimension & 2048 \\
CCA query heads & 8 \\
KV heads & 2 \\
Head dimension & 128 \\
Attention variant & CCGQA with CCA preconditioner \\
Query compression & $2\times$ \\
KV-cache compression & $8\times$ relative to full multi-head attention \\
Experts per MoE layer & 16 \\
Routing & Top-1, no residual expert \\
Expert FFN width & 4096 pre-activation / 2048 post-activation \\
Router latent dimension & 256 \\
Position embeddings & 50\% RoPE on each head \\
Tokenizer & Gemma3 tokenizer, 262{,}272 vocabulary size \\
Primary training hardware & AMD MI300X with Pollara networking \\
\bottomrule
\end{tabular}
\caption{ZAYA1-8B model configuration. Exact parameter counts are shown; the rounded release convention refers to the model as 0.7B active and 8B total. Architectural constants follow the ZAYA1 base configuration used for pretraining and continued post-training.}
\label{tab:model-config}
\end{table*}

\subsection{Architecture}

ZAYA1-8B uses an MoE architecture with three changes relative to contemporary MoE models: (1) CCA for the attention block, (2) the ZAYA1 router, and (3) residual scaling. In our ablations, these changes improve per-parameter perplexity relative to classical MoE architectures \citep{shazeer2016outrageously,fedus2022switch} using MLA or GQA attention and a linear router \citep{DeepSeekmoe}. CCA also improves training speed relative to GQA and MLA and reduces prefill FLOPs while maintaining comparable KV-cache compression rates.

\begin{figure*}
    \centering
    \includegraphics[width=0.6\linewidth]{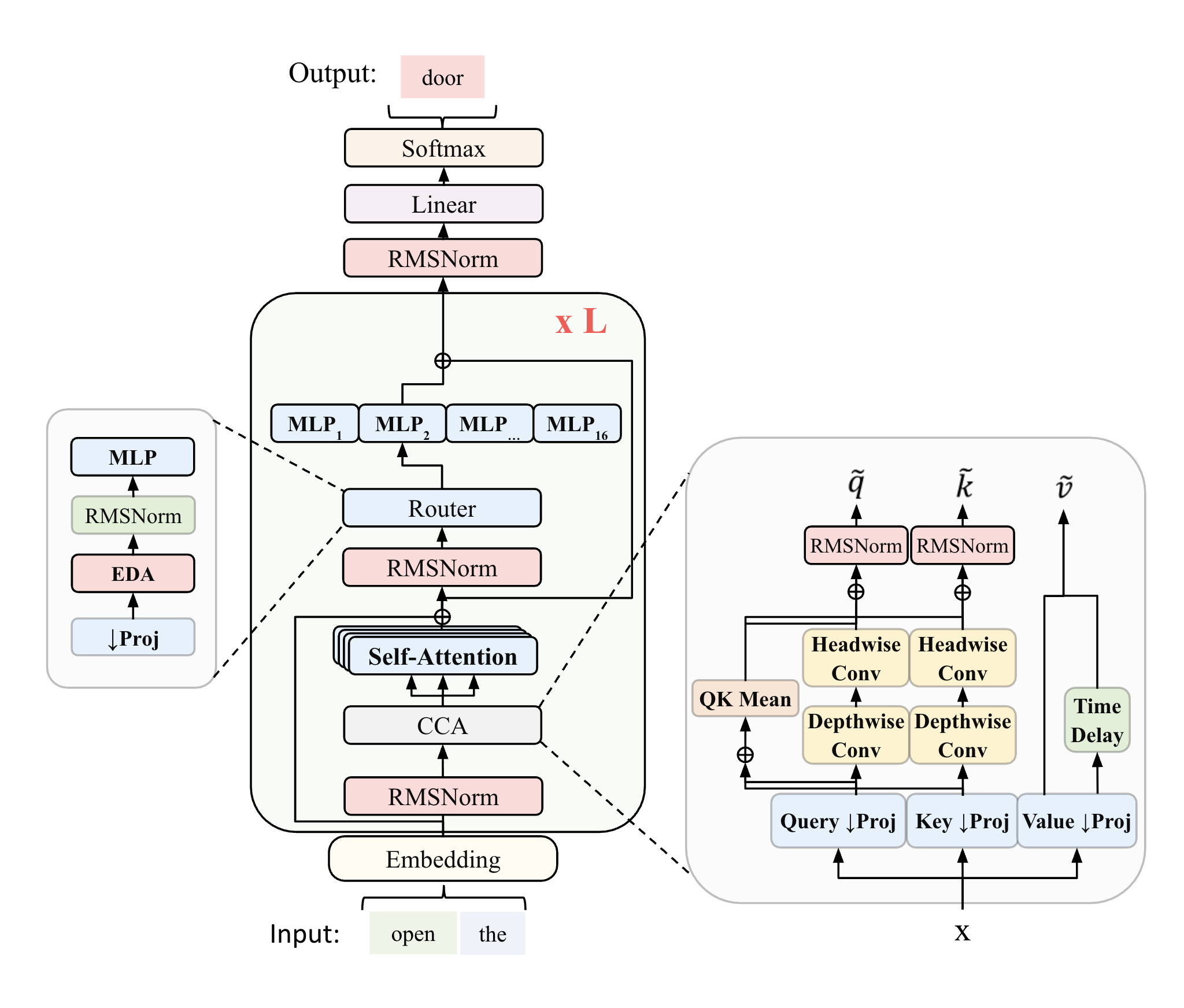}
    \caption{ZAYA1-8B model architecture. Two of the three main architectural changes are shown here: CCA for the attention block and the ZAYA1 router. The ZAYA1 router replaces the linear router with an MLP-based router consisting of a down-projection, EDA, and a three-layer MLP.}
    \label{fig:zaya_arch}
\end{figure*}

\subsubsection{Compressed Convolutional Attention (CCA)}
CCA performs sequence mixing in a compressed latent space using a lightweight convolutional downprojector. This reduces compute requirements for training and prefill and reduces KV-cache size for long-context decoding. CCA is competitive with attention variants such as MLA and GQA \citep{ainslie2023gqa,DeepSeekv3}. ZAYA1-8B's reasoning and long-context performance provides evidence that CCA remains effective at this scale and can support reasoning, in-context learning (ICL), and long-range recall. CCA also supports our long-context midtraining workloads at lower compute and communication cost, which was important for training ZAYA1-8B during midtraining and RL phases. Appendix~\ref{app:cca} provides additional details.

\subsubsection{ZAYA1 Router}
We replace the standard linear router used in many large-scale MoE models with a more expressive router. First, we use an MLP in place of the linear router. Second, we mix the router representation with the previous layer's routing representation using \textit{Exponential Depth Averaging (EDA)}, a variant of Depth-Weighted Averaging \citep{pagliardini2024denseformer}.

Given the residual stream input $x_l \in \mathbb{R}^{B \times S \times D}$, where $D$ is the residual stream dimension, the ZAYA1 router first down-projects the residual stream to a smaller router dimension $R$ using a learned weight matrix $W_{\text{down}} \in \mathbb{R}^{R \times D}$:
\begin{align}
    r_l = W_{\text{down}} x_l\,,
\end{align}
such that $r_l \in \mathbb{R}^{B \times S \times R}$. For ZAYA1-8B we set $R=256$.
We then apply EDA, which combines the representation with that of the previous layer using a learned coefficient $\gamma$:
\begin{align}
    r_l = r_l + \gamma r_{l-1}\,.
\end{align}
The EDA operation is followed by a three-layer MLP with GeLU activations to produce the final router scores $s \in \mathbb{R}^{B \times S \times E}$, where $E$ is the number of experts:
\begin{align}
    s_l = \text{softmax}(\text{MLP}(\text{RMSnorm}(r_l)))\,.
\end{align}
The scores are then used to select experts through a top-k operation:
\begin{align}
\label{eq:topk}
    e_{\text{idx}} = \text{topk}(s_l + b_l)\,,
\end{align}
where $b_l$ are learned bias-balancing vectors and topk selects the $k$ experts with the largest biased router scores for each token. In ZAYA1-8B, $k=1$, so \eqref{eq:topk} reduces to selecting $\operatorname*{arg\,max}_e( s_{l, e} + b_{l, e})$ for each token. The ZAYA1 router uses a bias-balancing scheme building on \citep{DeepSeekv3}. Routing biases are updated using a scheme inspired by proportional--integral--derivative (PID) controllers from classical control theory \citep{aastrom2006pid}.  The router enforces balancing across a global batch of expert choices. Our PID optimizer uses AdamW internally, where the error signal passed to the optimizer is the difference between the empirical routing probability distribution and the uniform distribution. Specifically, the gradient $\nabla b_{l, e}$, for expert $e$ at layer $l$, is computed as:
\begin{align}
\label{eq:bias_grad}
\nabla b_{l, e} = p_{l, e} - \frac{1}{E}\,,
\end{align}
where $p_{l, e}$ is the actual fraction of tokens routed to expert $e$ in the current batch, and $E$ is the total number of experts. This gradient signal is then used by AdamW to update the bias terms, penalizing over-utilized experts and boosting under-utilized ones. This improved the convergence speed and stability of the PID loop relative to the classical DeepSeek implementation.

\begin{figure}
    \centering
    \includegraphics[width=0.95\linewidth]{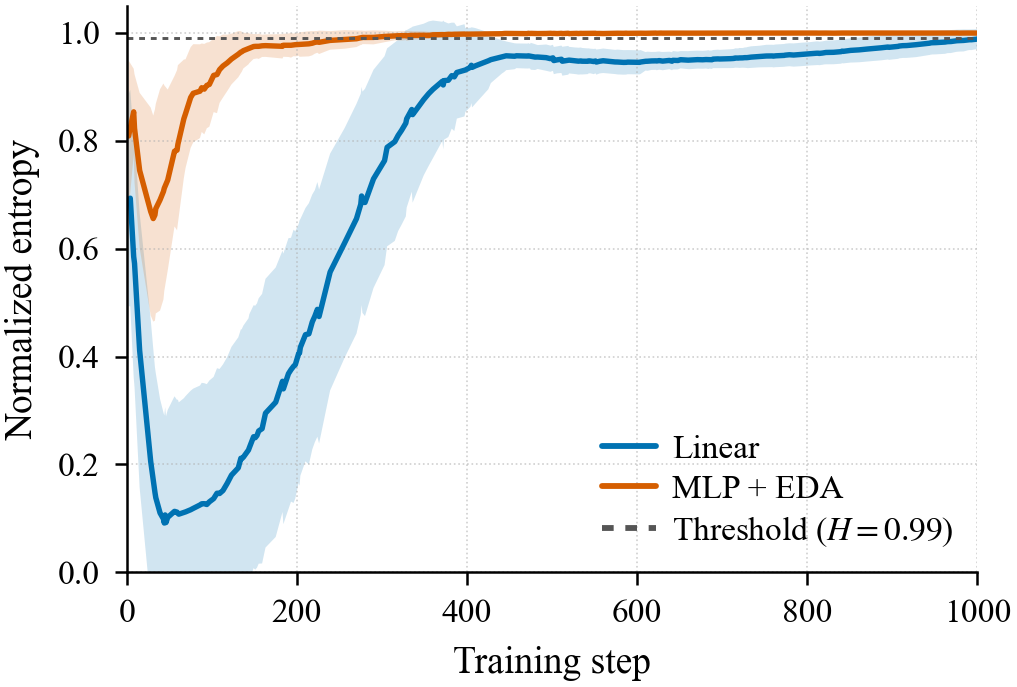}
    \caption{Normalized router-load entropy, averaged over MoE layers, as a function of training step from initialization of a 1.8B-parameter experimental model. For each global batch and layer, let $p_i$ denote the fraction of routed tokens assigned to expert $i$, with $E$ total experts. We report $H(p)/ \ln E$, where $H(p)=-\sum_{i=1}^E p_i \ln p_i$ is the Shannon entropy of the empirical expert-load distribution.}
    \label{fig:entropy}
\end{figure}

In our experiments, the MLP router and EDA improve MoE performance and make balancing (Figure \ref{fig:entropy}) and expert specialization easier. The additional MLP adds some FLOPs and parameters, but parameter-matched ablations show that the router is a strong target for marginal parameters compared with the experts or attention. The added router parameters and FLOPs remain small because the MLP operates in the down-projected latent space rather than in the full embedding dimension. Figure \ref{fig:entropy} illustrates the average balancing across layers from initialization of an experiment-sized model. Empirically, reduced time to convergence translated to increased recovery speed in the face of perturbations such as data distribution shifts throughout phases of training. This yields an improved router-load entropy convergence in the reported 1.8B ablation and reduced balancing failures in our training runs compared to linear routers.

\subsubsection{ZAYA1 Residual Scaling}
The final architectural change in ZAYA1-8B is residual scaling. We apply a learned bias $b_l$ and gating coefficient $\alpha \in \mathbb{R}^{D}$ both to the residual stream and to the output of each layer before the residual connection:

\begin{align}
    \text{Res-scale}(x) &= \alpha x + \beta , \\
    x_{l+1} = \text{Res-scale}_{res}(x_l) 
    &+ \text{Res-scale}_{out}\left(\text{Layer}(\text{RMSnorm} \nonumber
    (x_l))\right) 
\end{align}

Different gating coefficients and biases are applied to the residual stream and to the layer outputs. Residual scaling lets the model downweight parts of the residual stream and control how much prior residual information is retained. In our experiments, residual scaling provides similar benefits to Qwen's attention gating scheme \citep{qiu2025gated}, without the parameter or FLOP overhead of an explicit gating matrix. Residual scaling also helps control residual-norm growth through network depth, without observing any gradient vanishing. We initialize $\alpha$ to ones and $\beta$ to zeros, as this initializes the model with default residual connections. Because residual scaling adds only $4 \times L \times D$ parameters, its parameter and FLOP overhead are comparable to LayerNorm and are negligible.

Beyond these architectural changes, we trained with 16 experts and a hidden-dimension expansion factor of 2. This relatively fine-grained expert configuration improved performance at fixed parameter count, consistent with prior work \citep{team2025kimi,DeepSeekai2024DeepSeekv32,DeepSeekmoe,tian2025towards}.

Unlike many contemporary MoEs, we trained with top-k equal to 1 and without residual experts \citep{rajbhandari2022deepspeed,DeepSeekv3}. In our experiments, the improved routing expressiveness of the ZAYA1 router and the resulting expert specialization make a residual expert unnecessary. FLOP-matched experiments also favored top-1 over higher top-k when using the ZAYA1 router. We hypothesize that the ZAYA1 router assigns more certain expert choices, with better expert specialization, so additional experts in parallel via top-k are less useful. When larger values of $k$ are used, their contribution is further reduced by multiplication with the routing probability. ZAYA1-8B produces lower-entropy routing probabilities per token than linear routers, consistent with more confident routing. As a sanity check on expert redundancy, Appendix~\ref{app:expert-subspace-overlap} measures within-layer expert subspace overlap for ZAYA1-8B and public MoE baselines. ZAYA1-8B is not an outlier toward higher expert overlap: its first-projection input overlap is $1.45\times$ the random-subspace baseline, close to Qwen3-30B-A3B's $1.48\times$, while its output-projection overlap is intermediate among the compared MoEs. For attention, we used CCGQA with a query compression rate of $2\times$ and a KV compression rate of $8\times$. We applied RoPE \citep{su2023rotary} to half the channels in each head, leaving the other half without position embeddings. ZAYA1-8B was trained with the Gemma3 tokenizer.

Table~\ref{tab:model-config} summarizes core architectural hyperparameters of the final release configuration.

\section{Pretraining and Midtraining}
\label{sec:pretraining}

ZAYA1-8B was initialized from Zyphra's ZAYA1 base architecture and trained through pretraining, context-extension midtraining, and SFT on an AMD MI300X cluster equipped with the AMD Pensando Pollara networking stack. Full details of the base-model pretraining system, hardware, checkpointing, context parallelism, and AMD-specific optimizer and kernel work are provided in \citep{prior-zyphra-work}.

Table~\ref{tab:training-recipe} summarizes the main phases. Base pretraining used a broad web-crawl distribution with code, math, multilingual, and reasoning data mixed in progressively. The second base pretraining phase upweighted code, math, reasoning, and instruction-formatted data while still training at 4K context length. We then ran a reasoning-focused midtrain phase at 32K context for 1.2T tokens at a RoPE base frequency of 1M. This was followed by an SFT phase at 131K context for 660B tokens at a RoPE base frequency of 5M. We believe that training for a large number of tokens at longer contexts significantly improves the model's native long-context capabilities and thus provides a stronger base for post-training and RL. The substantial reduction in prefill FLOPs we obtained through using CCA was instrumental in making this feasible at our compute scale.

\begin{table*}[h]
\centering
\small
\begin{tabular}{lllll}
\toprule
Phase & Context & RoPE base & Token budget & Main emphasis \\
\midrule
Base pretraining, phase 1 & 4K & 10K & 8T & Broad web, code, math, multilingual \\
Base pretraining, phase 2 & 4K & 10K & 4T & More code, math, reasoning, instruction data \\
32K midtraining & 32K & 1M & 1.2T & Long-CoT reasoning, code, math, long-context data \\
SFT & 131K & 5M & 660B & Chat template, reasoning, code, IF, TTC traces \\
\bottomrule
\end{tabular}
\caption{Training recipe summary. Base-pretraining details are summarized here for context and described in detail in \citep{prior-zyphra-work}. }
\label{tab:training-recipe}
\end{table*}

Table~\ref{tab:midtrain-mixes} reports coarse data categories for the reasoning-focused midtrain and SFT. Percentages are normalized over the nonzero mixture weights in the data cards; we report only category-level proportions rather than individual dataset names. To specialize the model for reasoning and provide as strong a base for RL as possible we utilized a very high fraction of long-CoT reasoning traces in the midtrain and SFT. 

\begin{table*}[h]
\centering
\small
\begin{tabular}{lcc}
\toprule
Category & 32K midtraining & 131K SFT \\
\midrule
Long-CoT reasoning traces & 86.1\% & 75.0\% \\
Web, synthetic web, multilingual & 5.7\% & 9.8\% \\
Natively long-context data & 0.8\% & 6.4\% \\
Code corpus / code SFT & 3.0\% & 5.0\% \\
Math/STEM corpus & 3.0\% & 2.6\% \\
Short instruction / few-shot data & 1.4\% & 1.2\% \\
\bottomrule
\end{tabular}
\caption{Coarse midtraining data mixtures. The 32K context-extension mixture was trained for approximately 1.2T tokens, while SFT was trained for approximately 660B tokens; percentages denote normalized mixture weights. Individual source datasets are omitted.}
\label{tab:midtrain-mixes}
\end{table*}

For context extension, we used all-gather KV context parallelism with two ranks at 32K and eight ranks at 131K. CCA's compressed KV representation kept activation and KV-cache memory overhead low, while short asynchronous point-to-point exchanges handled the convolution and value-shift boundary conditions introduced by CCA. Across these phases, we trained with the Muon optimizer using AdamW RMS matching \citep{jordan6muon, moonshot-muon}.

\subsection{Reasoning-aware pretraining and answer-preserving trimming}
\label{sec:ap-trimming}

Recent work suggests that introducing long chain-of-thought reasoning data during pretraining and midtraining, rather than only during post-training, can produce gains that subsequent fine-tuning does not recover \citep{nvidia-synergy}. We follow this approach throughout ZAYA1-8B's training pipeline: every pretraining and midtraining phase included long-CoT data and it was a majority of the mix for the midtraining phases.

Including reasoning data at short pretraining contexts creates a practical challenge: reasoning traces from strong teacher models often exceed 10K tokens, with a long tail beyond 30K. At the initial 4K context length, each example must be handled in one of three ways: (i) drop it entirely, losing the reasoning signal; (ii) truncate naively, often preserving the reasoning prefix while losing the answer and thereby training the model on reasoning that never reaches a conclusion; or (iii) preserve the answer while truncating part of the reasoning. We use the third option and call the resulting scheme \emph{answer-preserving (AP) trimming}.

Given a sample containing one or more assistant messages with \texttt{<think>...</think>} reasoning blocks followed by a final-answer section, AP-trimming applies the following procedure to fit the sample within a target context budget $C$:

\begin{enumerate}[leftmargin=*,itemsep=2pt,topsep=2pt]
    \item \textbf{Keep unchanged.} If the full conversation fits within $C$, retain it as-is.
    \item \textbf{Trim the tail of the last reasoning block.} If the conversation does not fit, truncate the final assistant turn's reasoning trace from the tail, immediately before the answer. This preserves the start of the reasoning trace and the full answer section. The retained reasoning length is chosen so that the full sample fits within $C$.
    \item \textbf{Drop prior reasoning blocks.} For multi-turn conversations, if step~2 is insufficient, remove the \texttt{<think>} blocks of earlier assistant turns while preserving their answer sections, then re-apply step~2.
    \item \textbf{Drop the sample.} If the answer sections alone exceed $C$, discard the sample.
\end{enumerate}

The core idea is to truncate from the tail of the reasoning trace rather than from the middle. The beginning of a reasoning trace often contains problem decomposition, planning, and exploration of multiple approaches. The tail is usually more local, consolidating the selected approach into the final answer. Removing tail tokens therefore preserves more of the planning and decomposition signal while producing partial but coherent reasoning sequences whose beginning, truncated end, and final answer remain causally aligned. The transition between truncated reasoning and the answer is distributionally artificial, but in practice we did not observe obvious artifacts: pass-rate evaluations on reasoning benchmarks after pretraining and midtraining remained strong, and we did not identify a truncation-specific failure mode in downstream evaluations.

\paragraph{Stage-aware re-trimming} AP-trimming is applied offline to each dataset at each context length where the data is used. As the training pipeline advances through 4K pretraining, 32K midtraining, and 131K context-extension SFT, we re-trim each dataset to the corresponding context length and progressively retain longer reasoning traces. Most reasoning datasets fit fully at 131K context, so late midtraining operates on near-complete traces; early pretraining uses the most aggressive trimming.

\paragraph{Relation to prior work} The closest related techniques operate during inference or RL rollout generation rather than during pretraining data construction. \citep{scalerl} use forced length interruptions during RL rollouts: when a thinking trace approaches the budget, the environment appends an end-of-thinking phrase that forces the model to produce a final answer. \citep{yang2025qwen3} use a similar mechanism for inference-time thinking-budget control. Both methods operate on rollouts during training or generation, not on training data before consumption. The closest training-data analogue is the answer-length-filtered subset of \citep{nvidia-synergy}, which retains examples whose answer length exceeds 4K tokens as a proxy for reasoning depth. That is a selection strategy rather than a truncation strategy. AP-trimming addresses the complementary problem of using long-CoT reasoning data at training contexts shorter than the natural trace length by truncating reasoning while preserving the answer section.

\section{Post-training}
\label{sec:posttraining}

\begin{figure*}
    \centering
    \includegraphics[width=\linewidth]{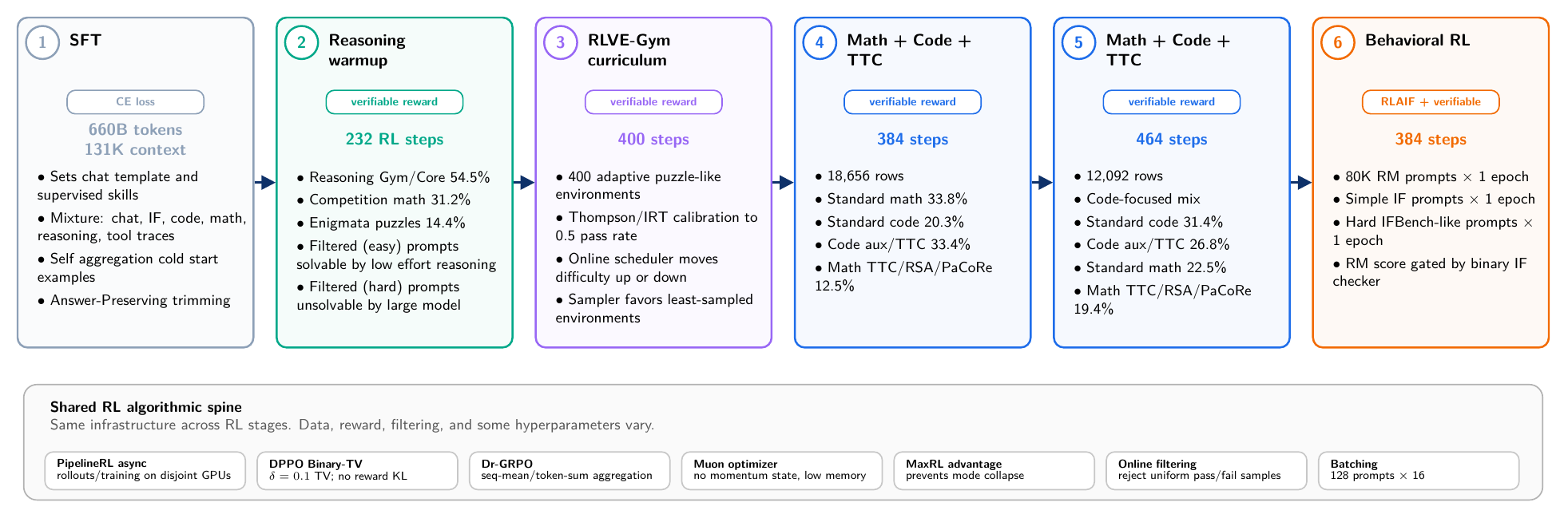}
    \caption{Schematic of our post-training process for ZAYA1-8B. Post-training progressed through SFT followed by four sequential RL stages. The first stage built general reasoning capabilities on math and puzzles and was then followed by two stages of code + TTC training. The model was then polished through a short behavioral RLHF phase which focused more on chat and user interaction.}
    \label{RL_cascade_fig}
\end{figure*}

Post-training begins with SFT, followed by a four-stage RL cascade. The first three RL stages are almost entirely verifiable reasoning: a math/puzzle/TTC warmup, an RLVE-Gym adaptive difficulty curriculum, and a two-phase math+code+TTC stage. We defer general chat, style, and instruction-following optimization to the final behavioral RL stage. This ordering prioritizes capability extraction from verifiable signals before applying preference and instruction-following rewards.

Two aspects of this ordering differ from common post-training recipes. First, reasoning RL is front-loaded: most RL compute before behavioral RL is spent on verifiable math, puzzles, synthetic environments, and code. Second, the code stage uses several synthetic auxiliary environments constructed from competitive-programming references, including code input/output prediction, code reconstruction from test cases, and falsification.

\subsection{Supervised Fine-Tuning}
\label{sec:sft}

The SFT phase establishes the chat template used in subsequent post-training, improves instruction following, and continues reasoning supervision at 131K context. The stage consumed 660B tokens. We use a supervised mixture spanning chat, instruction following, code, math, reasoning, tool-calling traces, and TTC aggregation examples, but do not report individual dataset details.

Because the SFT stage trains at 131K context, packing strategy mattered. We use optimized best-fit decreasing bin packing \citep{ding2024fewertruncationsimprovelanguage} rather than naively streaming examples into fixed-length windows and truncating at arbitrary boundaries. The packer fills each 131K window with complete examples whenever possible; over-length examples are handled by dataset-specific preprocessing before packing rather than by training on arbitrary suffixes created by a fixed-boundary truncation pass. This avoided hallucination artifacts we observed when models were trained on endings of mechanically truncated packed sequences.

SFT also introduces aggregation-based examples used by Markovian RSA. These examples present the model with a problem and several candidate reasoning tails, then train it to produce a single improved solution. Section~\ref{sec:ttc-training} describes this construction in detail.

\subsection{Reinforcement Learning Cascade}
\label{sec:rl-cascade}

Post-training reinforcement learning is organized as a four-stage cascade. The cascade uses a shared algorithmic spine described in Section~\ref{sec:rl-algo}; individual stages differ in data, reward signal, and stage length.

\begin{table*}[h]
\centering
\small
\begin{tabular}{llll}
\toprule
Stage / phase & Length & Main data & Reward signal \\
\midrule
Reasoning warmup & 232 steps & Math, puzzles, TTC traces & Verifiable task reward \\
RLVE-Gym curriculum & 400 steps & 400 adaptive environments & Environment verifier / solve rate \\
Math+Code+TTC, phase 1 & 384 steps & General math, code, TTC & Verifiable task reward \\
Math+Code+TTC, phase 2 & 464 steps & Code-focused mix & Verifiable task reward \\
Behavioral RL & 384 steps & Chat/RM, simple IF, hard IF & RM score with IF gate for IF stages \\
\bottomrule
\end{tabular}
\caption{RL cascade summary. The first three stages emphasize verifiable reasoning. Behavioral RL is run last to tune chat, style, and instruction-following behavior.}
\label{tab:rl-cascade-summary}
\end{table*}

\subsubsection{Algorithmic Spine}
\label{sec:rl-algo}

All RL stages and subphases share a common algorithmic spine. Per-stage differences are confined to data, reward signal, and a small number of hyperparameters.

\paragraph{PipelineRL} Rollout generation and gradient updates run fully asynchronously on disjoint GPU pools \citep{pipelinerl, scalerl}. We allocate 2--5$\times$ more rollout workers than trainer workers, balanced per workload to match average response length to actor update time so that neither pool stalls. Trainer-to-rollout weight sync happens in place every 2 trainer iterations; in steady state, the rollout policy is bounded at 2 trainer updates behind the trainer policy.

\paragraph{Trust region} DPPO Binary-TV. We replace PPO's per-token ratio clipping with Binary Total-Variation trust-region masking \citep{dppo}. Tokens for which the policy-divergence estimate exceeds a threshold $\delta$ are masked from the gradient while remaining tokens contribute as in standard policy-gradient updates. We use $\delta = 0.1$ in production. We tune this threshold against preserving the reward-growth trajectory of an unconstrained baseline, selecting the largest value that did not produce unconstrained reward growth in early training. The Binary-TV variant uses a deterministic indicator over a single divergence threshold rather than the continuous TV penalty or the Top-$K$ approximation, and adds negligible overhead relative to standard PPO.

\paragraph{Loss aggregation} Dr-GRPO SMTSN. Loss aggregation follows Dr-GRPO \citep{dr-grpo}: sequence-mean over token-sum-norm (SMTSN). Token-level losses are summed within each rollout and then averaged across rollouts in the batch, rather than averaged per-token. This avoids the implicit length normalization in standard GRPO, which biases the gradient toward longer responses.

\paragraph{Advantage estimation} MaxRL. Advantages are computed as in \citep{tajwar2026maxrl}. For each prompt, we sample a group of $G$ rollouts with task rewards $r_i \in \{0,1\}$ using dynamic sampling \citep{dapo}. The advantage normalizes by the per-prompt mean reward rather than the per-prompt reward standard deviation:
\begin{equation}
\hat{A}_i = \frac{r_i - \bar{r}}{\bar{r}},
\end{equation}
where $\bar{r} = \tfrac{1}{G}\sum_{j=1}^{G} r_j$ is the group mean reward. This corresponds to the variance-reduced MaxRL estimator (\citet{tajwar2026maxrl}, Algorithm~1), which is unbiased for a truncated maximum-likelihood objective rather than for expected reward and produces stronger gradient signal on harder prompts. We use this normalization in all RL stages except the final behavioral RL stage (Section~\ref{sec:rl-behavioral}), which uses standard GRPO with reward standard-deviation normalization.

\paragraph{Reward shape and length reward} Task rewards are binary across the cascade, with the exception of (i) RLVE-Gym environments that yield continuous solve rates near difficulty thresholds (Section~\ref{sec:rl-rlve}) and (ii) the behavioral RL stage, which uses a normalized reward-model score (Section~\ref{sec:rl-behavioral}). All RL stages except behavioral RL also include the difficulty-scaled length reward of Section~\ref{sec:rl-token-efficiency}, applied additively as $\Delta r_i$ to the task reward at the numerator of the advantage only --- the denominator $\bar{r}$ uses the unmodified task reward to avoid scale blow-up. The length-reward coefficient $c$ ramps from a small initial value during reasoning warmup to $c=1.0$ during the math+code+TTC stage, where the production reasoning length is established.

\paragraph{No KL in reward} The cascade applies no KL regularization to the reward; the trust region is enforced entirely by DPPO Binary-TV. In stress-testing with high KL-penalty coefficients, we observed a length-dependent bias attributable to applying a signed sequence-level log-ratio reward term to stale or mixed-policy rollouts under in-flight weight sync; Section~\ref{sec:stability} describes the mechanism and possible mitigations. The production cascade avoids this configuration entirely by relying on the DPPO trust region alone.

\paragraph{Optimizer} All RL stages use Muon with momentum set to zero \citep{glm5team2026glm5vibecodingagentic}, extending the GLM-5 prescription of resetting the optimizer at each weight-sync boundary into a fully momentum-free regime. Section~\ref{sec:optimizer} describes this choice in detail and discusses its motivation and memory implications.

\paragraph{Hyperparameters} Across all five stages, the cascade uses minibatches of 128 prompts with rollout group size $G=16$ responses per prompt. Per-rollout maximum response length is 81{,}920 tokens, except for the first half of the reasoning-warmup stage, which uses 65K. The maximum aggregation-prompt length is 20{,}480 tokens, sized to fit Markovian RSA round-1 prompts containing $C=4$ candidate tails. Trainer-to-rollout weight sync occurs every 2 trainer iterations. The trainer requires 2 batches of completed rollouts to be available in the buffer before pulling, and the buffer is capped with oldest-sample eviction; in steady state, on-policy staleness is bounded at 2 trainer updates. Learning rates are set per stage in the range $2 \times 10^{-6}$ to $1 \times 10^{-5}$, with the smallest values used during behavioral RL.

\subsubsection{Token efficiency}
\label{sec:rl-token-efficiency}

To encourage concise reasoning, we combine aspects from ALP~\citep{xiang2025alp} and ShortRL~\citep{yuan2025shortrl} to create a group-relative, difficulty-scaled length reward.

Given a prompt with rollout group size $G$, response reward $r_i \in \{0,1\}$, and response length $\ell_i$, we compute the group solve rate $p = \frac{1}{G}\sum_{i=1}^{G} r_i$ and the shortest correct response length $\ell_{\min}$. Similar to ShortRL, we define a linear length interpolation, with the distinction that $\ell_{\max}$ is a constant:

\begin{equation}
\begin{aligned}
\lambda_i &= \frac{1}{2} - \operatorname{clamp}\!\left(\frac{\ell_i - \ell_{\min}}{\ell_{\max} - \ell_{\min}},\; 0,\; 1\right)\,,\\
\tilde{\lambda}_i &=
\begin{cases}
\frac{1}{2} & \text{if } \ell_i \leq \ell_{\min} + T_\ell\, , \\
\lambda_i & \text{otherwise}\, .
\end{cases}
\end{aligned}
\end{equation}

Let $k=\sum_i r_i$ denote the number of correct responses in the group. We apply the length reward only when at least two responses in the group are correct, so that there is a nontrivial comparison among correct response lengths. We adopt the following correctness and difficulty gate:

\begin{equation}
m_i = \mathbbm{1}\!\left[p > p^* - T_{\text{acc}}\right] \cdot \mathbbm{1}\!\left[p > \tfrac{1}{G}\right] \cdot \mathbbm{1}\!\left[r_i = 1\right]\,,
\end{equation}

\noindent where the term $p^*$ denotes a running maximum solve rate for the corresponding data source or environment, the condition $p > 1/G$ is equivalent to $k \geq 2$ for integer-valued binary rewards, and $T_{\mathrm{acc}}$ is a tolerance that prevents the length reward from activating far below the current observed capability frontier. We additionally scale the bonus by the solve rate $p$, attenuating the length penalty on difficult problems and amplifying it on easier problems. The final additive reward is:
\begin{equation}
\Delta r_i = c \cdot p \cdot m_i \cdot \tilde{\lambda}_i\, ,
\end{equation}
\noindent where $c$ is a scaling coefficient. This reward $\Delta r_i$ is added to the task reward, biasing the policy toward shorter correct solutions while preserving task accuracy in our production runs.

\subsubsection{Reasoning Warmup}
\label{sec:rl-warmup}

The first RL stage is a 232-step reasoning warmup on math, puzzle, and TTC reasoning prompts. Its purpose is to adapt the SFT model to long verifiable rollouts before the broader RLVE and math+code stages. The warmup set contains 84{,}604 rows and is deliberately biased toward hard prompts: retained examples have prior pass rate at most 0.75, with most examples at substantially lower pass rates. Responses in this stage are long, with median replay response length around 17.6K tokens and a p90 near 30K tokens.

\begin{table*}[h]
\centering
\small
\begin{tabular}{lr}
\toprule
Category & Mix share \\
\midrule
Reasoning-gym and reasoning-core puzzles & 54.4\% \\
Competition level math reasoning & 31.2\% \\
Enigmata puzzles & 14.4\% \\
\bottomrule
\end{tabular}
\caption{Coarse composition of the reasoning-warmup RL data. Percentages are computed over 84{,}604 warmup rows.}
\label{tab:rl-warmup-mix}
\end{table*}

Rewards are verifiable and task-specific. For math problems, the reward is based on final-answer correctness after normalization. For puzzle environments, the reward is supplied by the environment verifier. TTC prompts are formatted to match the Markovian RSA workflow described in Section~\ref{sec:ttc}, so the model begins RL already seeing aggregation-based reasoning prompts.

\subsubsection{RLVE-Gym Difficulty Curriculum}
\label{sec:rl-rlve}

The second RL stage trains for 400 steps on 400 adaptive and verifiable problem generators from RLVE-Gym~\citep{zeng2025rlve}. We integrated RLVE as a dataset in VeRL~\citep{sheng2025hybridflow}. Although this stage has fewer optimizer steps than the later math+code stage, the average step is roughly twice as long because responses are long, with reasoning lengths around 50K tokens. We use this stage to expose the model to a broad distribution of puzzle-like verifiable environments while keeping each environment near the model's current difficulty boundary.

During training, we used an online scheduler for problem difficulty, and we balanced environment selection using a weighted sampler for which the least sampled environments get the highest weight. Our difficulty scheduler differs slightly from the authors' in that it uses a tighter bound on the difficulty $d$ and allows regressions.

Let $\hat{y}$ denote either the optimal solution or a reasonable heuristic when optimal solutions are intractable. We define
\begin{equation}
\begin{aligned}
r_i &=
\begin{cases}
1 & \text{if } |\hat{y}-y| < \epsilon, \\
0 & \text{otherwise,}
\end{cases}
\\[0.5em]
\delta_i &=
\begin{cases}
+1 & \text{if } \bar{r} > 0.7 \text{ and } d_{\mathrm{group}} = d, \\
-1 & \text{if } \bar{r} = 0, \\
0  & \text{otherwise,}
\end{cases}
\\[0.5em]
d &\leftarrow d + \delta_i ,
\end{aligned}
\end{equation}

where $r_i$ is reward per rollout in a group, $\bar{r}=\tfrac{1}{G}\sum_i r_i$ is the group pass rate, $\epsilon$ is an environment-specific numerical tolerance used to determine whether the rollout answer $y$ is close enough to the target $\hat{y}$ to receive reward 1, $d$ is the current difficulty setting, and $d_{\text{group}}$ is the observed difficulty of the last computed group. We constrain updates to $d_{\text{group}} = d$ to prevent stale difficulties from affecting the pass rate estimate.

Crucially, we use an initial tuning step to avoid training on difficulties that are too easy for the model, and we aim to maximize the information content during training by initializing all environments to a difficulty that gives a 0.5 pass rate for the model. This tuning process is an adaptive search problem, and the search space is essentially unbounded. Some environments are solvable into the range of $d > 100$, while others are rarely solvable even at 0. For this reason, we rely on Thompson Sampling~\citep{thompson1933likelihood} as a reasonably efficient method to determine the 0.5 solve rate crossing point for every environment. We model the pass rate using the complement of the logistic curve as is commonly done in Item-Response-Theory (IRT)~\citep{lord1980irt} and we sample from the midpoint with an $\varepsilon$-greedy approach. Each verified response group yields an estimate of the pass rate. A parameter pool is maintained as with Thompson Sampling and a single Gaussian prior on $\mu$ and $s$ is used for all environments based on empirically observed ranges.

\begin{equation}
\begin{aligned}
    p_{\text{success}} &= \sigma\!\left(-\frac{d-\mu}{s}\right) = \frac{1}{1 + e^{(d-\mu)/s}}\,,\\
    \boldsymbol{\Theta} &= \{(\mu_m, s_m)\}_{m=1}^{M}\,.
\end{aligned}
\end{equation}

Given a parameter pool, we perform weighted sampling proportional to the posterior weight of the candidates (initialized as uniform) in $\Theta$ and for each iteration we sample a candidate and use it to compute a difficulty $d$ at $p_{\text{target}}$:

\begin{equation}
\begin{aligned}
    \boldsymbol{w} &= (w_1,\ldots,w_M), \qquad \sum_{m=1}^{M} w_m = 1\,,\\
    j &\sim \mathrm{Categorical}(\boldsymbol{w})\,,\\
    d_j &= \mu_j + s_j \cdot \log\left(\frac{1 - p_{\text{target}}}{p_{\text{target}}}\right)\,,
\end{aligned}
\end{equation}
where $\boldsymbol{w}$ is the normalized posterior-weight vector over the candidate logistic-curve parameters in $\boldsymbol{\Theta}$.

We use $p_{\text{target}} = 0.5$, the maximum Fisher information point of the logistic model~\citep{lord1980irt}, with $\varepsilon$-greedy exploration to $0.5 \pm 0.25$. We then perform rollouts and verification at the sampled difficulty and close the loop by updating and renormalizing the posterior:

\begin{equation}
\begin{aligned}
w_j &\propto
w_j \cdot \mathrm{Binomial}\!\left(k;\,G,\,p_{\mathrm{success},j}\right),
\\
&\hspace{2em} j \in \{1,\ldots,M\}.
\end{aligned}
\end{equation}

\noindent where $p_{\text{success},j}$ is the current estimate for candidate $j$. Groups are generated asynchronously using vLLM with the previous phase's frozen model weights. If the effective sample size of the pool falls below a threshold, we resample with replacement from $\Theta$, aggregate the observation history as a recency-weighted sum of successes and failures, then re-initialize likelihoods.

This curriculum is intended to maximize useful verifier signal. Environments that are too easy produce mostly positive groups and little policy-gradient information; environments that are too hard produce mostly negative groups. The initial calibration and online difficulty updates keep each environment near a solvable but non-saturated regime, making the stage a bridge between the narrower reasoning warmup and the broader math+code+TTC RL stage.

\medskip

\subsubsection{Math, Code, and Test-Time Compute RL}
\label{sec:rl-mathcodettc}

The third RL stage is the main capability-building stage of the cascade. It combines olympiad-level math, competitive-programming code, Markovian RSA aggregation prompts, PaCoRe continuation prompts, and synthetic auxiliary code environments. We run this stage in two phases: a 384-step general math+code+TTC phase, followed by a 464-step code-focused phase.

Table~\ref{tab:rl-math-code-mix} summarizes the two data mixtures. Phase 1 contains 18{,}656 rows and balances math and code while introducing TTC and PaCoRe variants. Phase 2 contains 12{,}092 rows and increases the code share while retaining math TTC data.

\begin{table*}[h]
\centering
\small
\begin{tabular}{lcc}
\toprule
Category & Phase 1: general & Phase 2: code-focused \\
\midrule
Standard code prompts & 20.3\% & 31.4\% \\
Code auxiliary / code TTC prompts & 33.4\% & 26.8\% \\
Standard math prompts & 33.8\% & 22.5\% \\
Math TTC / RSA / PaCoRe prompts & 12.5\% & 19.4\% \\
\bottomrule
\end{tabular}
\caption{Coarse composition of the math+code+TTC RL stage. Phase 1 uses 18{,}656 rows; phase 2 uses 12{,}092 rows. Percentages are grouped from source tags and rounded.}
\label{tab:rl-math-code-mix}
\end{table*}

The auxiliary code environments are constructed by transforming competitive-programming references into multiple verifiable tasks per source problem. Each seed problem contains a problem statement, input/output specification, accepted reference implementations, rejected or incorrect implementations when available, and test cases. From these seeds, we construct three auxiliary task families:

\begin{enumerate}[leftmargin=*,itemsep=2pt,topsep=2pt]
    \item \textbf{CodeI/O prediction}~\citep{li2025codeiocondensingreasoningpatterns}. Given code and a set of inputs, the model predicts the outputs; in the reverse direction, given code and outputs, the model proposes inputs that produce them. Output-prediction rewards use exact normalized agreement with the reference execution. Input-prediction rewards execute the reference program on the generated input and check that the target output is produced while satisfying the input schema.
    \item \textbf{CodeARC reconstruction}~\citep{wei2025codearcbenchmarkingreasoningcapabilities}. Given a problem description, input/output specification, and example test cases, the model synthesizes code. The verifier compiles or executes the generated solution and checks it against held-out tests.
    \item \textbf{Falsification.} Given a specification and a candidate implementation, the model must find an input that falsifies the implementation relative to the specification or a trusted correct implementation. The verifier checks that the generated input is valid and that it induces a disagreement or specification violation.
\end{enumerate}

These tasks target algorithmic reasoning primitives rather than only end-to-end competitive-programming solving. CodeI/O emphasizes execution tracing and inverse reasoning over program behavior. CodeARC emphasizes synthesis from sparse behavioral evidence. Falsification emphasizes adversarial test construction and spec-implementation comparison. All three are binary-verifiable and therefore fit the same RL objective as math and puzzle prompts.

TTC prompts are included in the same RL stream. For Markovian RSA examples, the prompt contains the original problem and a small set of candidate reasoning tails. The policy generates a single aggregated solution and receives the standard verifiable reward for the final answer or produced code. This lets the stage train both ordinary single-rollout problem solving and the aggregation workflow used at inference time.

\subsubsection{Agentic task scope}
\label{sec:rl-agentic-scope}

ZAYA1-8B does not include a dedicated multi-turn agentic RL stage in this release. We include some supervised agent, tool, and SWE traces during SFT, but the RL cascade is primarily optimized for verifiable reasoning, math, code, and instruction-following behavior. As a result, we expect agentic benchmarks such as BFCL-v4 and $\tau^2$ to lag models whose post-training explicitly emphasizes multi-turn tool use. Scaling agentic data and agentic RL is left for future releases.

\subsubsection{Behavioral RL}
\label{sec:rl-behavioral}

The final RL stage tunes general chat behavior, style, and instruction following after the verifiable-reasoning stages have established the model's math and code capabilities. Behavioral RL uses standard GRPO with reward standard-deviation normalization rather than the MaxRL normalization used in the verifiable-reasoning stages. It also does not use the length reward from Section~\ref{sec:rl-token-efficiency}.

We first train for one epoch on 80K behavioral prompts \citep{wang2024helpsteer2,wang2025helpsteer3preferenceopenhumanannotatedpreference}. This stage improves general response quality and chat behavior without changing the reasoning-focused data distribution of the earlier stages.

We then run two instruction-following stages, each for one epoch. The first uses simpler instruction-following prompts; the second uses more difficult IFBench-like prompts. For these IF stages, the reward is gated by a binary instruction-following checker. If the completion fails the IF gate, its reward is set to zero. If it passes the gate, the completion is scored by the reward model. This prevents the reward model from assigning positive reward to fluent responses that fail the explicit instruction constraints.

\subsection{RL Infrastructure}
\label{sec:rl-infra}

\paragraph{Router replay} The single most important MoE-specific change for RL stability is router replay: the trainer reuses the expert routing assignments produced by vLLM at rollout time during its own forward pass over the rollout, rather than recomputing routing decisions from scratch. Even with the precision settings in Section~\ref{sec:precision}, small numerical differences between the rollout engine and the trainer can produce different routing decisions for tokens near a router decision boundary. In an MoE with top-1 routing, a token routed to expert $e_{\rm inference}$ at rollout time but to a different expert $e_{\rm train} \neq e_{\rm inference}$ at gradient time produces different per-token logits, which corrupts the on-policy gradient. Router replay eliminates this source of mismatch: by pinning the trainer's expert selection to the rollout-time decision, enforcing $e_{\rm train} \equiv e_{\rm inference}$, the gradient is computed against the same expert sequence that produced the rollout. We discuss the SNR view of this mismatch in Section~\ref{sec:discussion-router-replay}.

In practice, vLLM writes per-token and per-layer expert assignment indices to a shared memory buffer during decode. The write is overlapped with decode work to avoid slowing rollout generation. Assignments are then packed alongside the rest of the rollout batch (token IDs, masks, etc.) when the batch is shipped to the trainer, so router replay introduces no separate transport step.

\paragraph{Memory and recompute strategy} For long-rollout training the dominant memory pressure comes from activations. We combine host-side activation offloading with gradient checkpointing: the hidden state tensors from each layer that autograd must retain for backward are temporarily offloaded to CPU memory during the forward pass, while checkpointed layer interiors discard their forward activations and reconstruct them by rerunning the layer forward during backward. This trades extra backward-time compute and host-to-device traffic for substantially lower peak GPU activation memory. In this configuration we use FSDP shard size 4 under the FSDP2 sharding strategy with sequence parallelism disabled; at this model size and per-rank rollout length, the extra cross-rank communication from sequence parallelism and ring attention is not worth the memory savings.

\paragraph{Packing and dynamic batching} We use sequence packing and variable length attention for the trainer. This allows the trainer to run with dynamic microbatch sizing: rather than fixing the number of rollouts per microbatch, we fix a token budget of 131{,}072 tokens per GPU per microbatch and pack rollouts into microbatches up to this budget. This avoids paying for the longest rollout in a fixed-rollout-count microbatch when most rollouts are shorter, and keeps GPU memory utilization stable across batches even when rollout-length distributions shift between training stages. We additionally rebalance pack assignments across GPUs so that microbatches on different ranks contain comparable token counts; without this, the slowest rank gates the entire step, since synchronous gradient accumulation must wait for all ranks to finish. With balancing, per-step variance across ranks is small enough that no rank consistently bottlenecks training.

\paragraph{Buffer management} The trainer pulls completed rollouts from a shared buffer with a maximum capacity bound and oldest-sample eviction. As described in Section~\ref{sec:rl-algo}, the trainer requires 2 batches of completed rollouts to be available before pulling. This combination keeps the rollout pool from running ahead of the trainer (which would inflate staleness and waste rollout compute), while ensuring the trainer is never blocked waiting for fresh rollouts under our 2--5$\times$ rollout-to-trainer ratio.

\subsection{Precision}
\label{sec:precision}

\begin{figure*}
    \centering
    \includegraphics[width=0.92\linewidth]{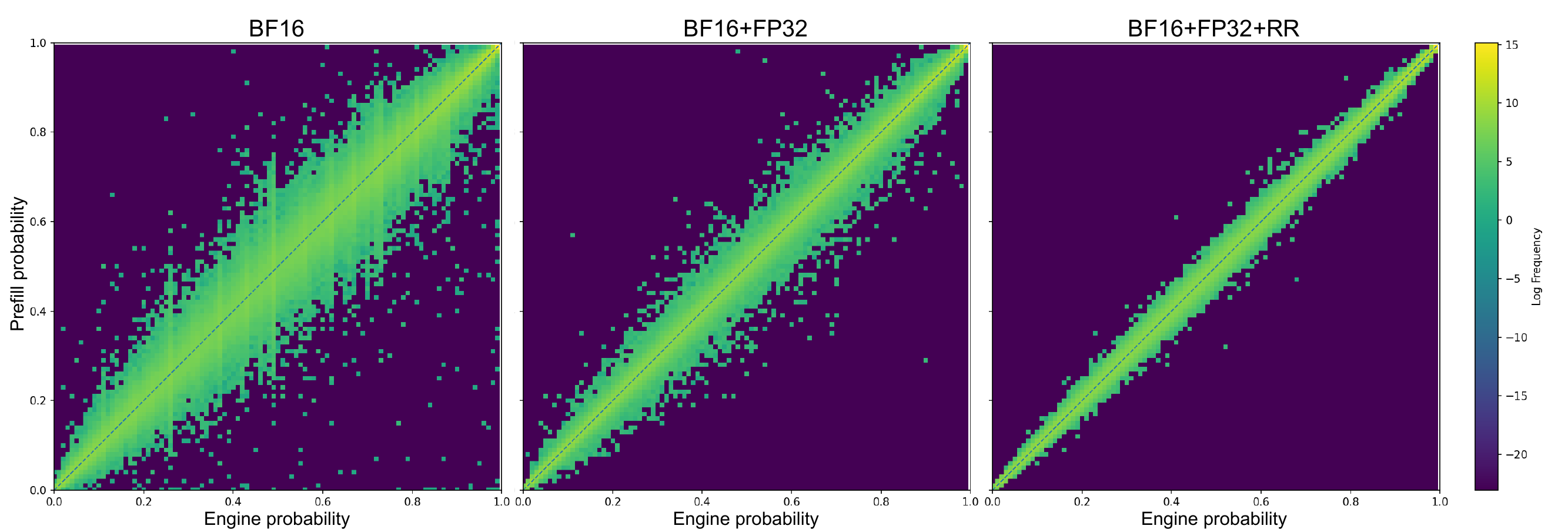}
    \caption{Per-token probability comparison (log scaled frequency): vLLM (engine, used for rollout generation) vs.\ trainer prefill (used for gradients) with incremental precision improvements. \textbf{BF16}: naive uniform BF16 implementation in inference and prefill. \textbf{BF16+FP32}: addition of selective upcasting of a subset of operations to FP32. \textbf{BF16+FP32+RR}: additional improvement from implementing router replay on trainer prefill from cached indices of rollout. Each point is a token from a 128-prompt, $G=16$ evaluation batch with 4K-token completions on ZAYA1-8B. Identity line shown (dashes). For BF16+FP32+RR, KL divergence $= 1.3 \times 10^{-4}$, Pearson $r > 0.9996$. }
    \label{fig:engine-trainer-match}
\end{figure*}

The default precision regime for ZAYA1-8B RL is BF16 weights and activations, with a small set of operations promoted to FP32. The subset of operations in FP32 is identical between the trainer and vLLM, which is necessary for engine-trainer log-prob agreement within the regime needed for stable PipelineRL training (see Figure~\ref{fig:engine-trainer-match}).

\paragraph{FP32 operation set} The following operations run with FP32 numerics on both trainer and inference paths:

\begin{itemize}[leftmargin=*,itemsep=2pt,topsep=2pt]
    \item \textbf{Loss/output:} fused cross-entropy accumulation and LM-head
    matmul.
    \item \textbf{Attention/normalization:} CCA cache state, QK-norm, QK-mean,
    and RMSNorm; see Section~\ref{sec:model}.
    \item \textbf{Routing/residuals:} router softmax and residual stream
    additions.
\end{itemize}
The LM-head FP32 promotion follows precedents in \citep{scalerl} and \citep{minimaxm1}. The remaining FP32 ops were added incrementally to close engine-trainer log-prob mismatch observed in early training runs; without them, mismatch produces grad-norm spikes and stale-policy artifacts under PipelineRL.

\paragraph{FP16 detour} Recent work argues that training--inference mismatch in RL fine-tuning can arise directly from floating-point precision, and proposes using FP16 uniformly rather than BF16 as a simple way to reduce mismatch \citep{qi2025fp16mismatch}. However, in our comparisons, we found that a hardened BF16 path with a small matched FP32 operation set on both the rollout engine and trainer achieved the engine--trainer agreement needed for stable PipelineRL training, while retaining BF16's dynamic-range advantages. We therefore use BF16 weights and activations by default, promote only the operations listed above to FP32, as described previously.

\paragraph{Rollout Engine-trainer match} Figure~\ref{fig:engine-trainer-match} compares per-token log-probabilities computed by vLLM (used during rollout generation) and by the trainer's prefill (used to compute gradients). At our default precision setup, the two distributions are nearly identical: KL divergence $= 1.3 \times 10^{-4}$ and Pearson $r > 0.9996$ over a 128-prompt, $G=16$ batch with 4K-token completions. This level of agreement is a precondition for stable PipelineRL training under our staleness regime; without the FP32 op set above, agreement degrades substantially and downstream training is unstable.

\subsection{Optimizer}
\label{sec:optimizer}

Let $\mathcal{L}_t(W)$ denote the actor training loss for the parameter matrix $W$ on rollout batch $t$, and let $g_t = \nabla_W \mathcal{L}_t(W)$ be the corresponding actor gradient. Let $m_t$ denote Muon's first moment buffer, and let $\mathcal{M}(\cdot)$ denote the Muon orthogonalization step via Newton-Schulz \citep{jordan6muon}. Standard Muon uses the update
\begin{equation}
\begin{aligned}
    m_t &= \mu m_{t-1} + g_t\,, \\
    \Delta W_t &= - \eta_t \mathcal{M}(m_t)\,,
\end{aligned}
\end{equation}
where $\eta_t$ is the learning rate at optimizer step $t$. For actor updates, we set $\mu=0$ so $m_t = g_t$ and $
\Delta W_t = - \eta_t \mathcal{M}(g_t)$. Thus each actor update depends on the current rollout batch and does not carry first-moment optimizer state across rollout batches. For embedding and output-head parameters, including the word embedding and LM head, we use AdamW rather than Muon. For the remaining matrix-valued actor weights, we use momentum-free Muon.

The motivation differs from pretraining. Compared to AdamW, Muon stands as a more compute efficient optimizer that is well suited to the RL setting where updates to parameters are sparse~\citep{mukherjee2026needadamsurprisinglystrong}. Furthermore, in next-token pretraining, adjacent minibatches are drawn from a comparatively stationary data distribution, so momentum can average compatible gradient directions across steps. In RL, each actor update is tied to a rollout batch whose prompts, sampled trajectories, rewards, and generating policy snapshot may differ from neighboring batches. Following~\citep{glm5team2026glm5vibecodingagentic}, we view optimizer-state reset as a useful stability heuristic for asynchronous RL. Our setting extends this idea: instead of resetting the optimizer state only at rollout-engine weight-sync boundaries, we make every actor update momentum-free. This makes each update depend only on the current rollout batch while retaining Muon's normalized matrix update, $\Delta W_t = - \eta_t \mathcal{M}(g_t)$, rather than a raw SGD step ($\Delta W_t = - \eta_t g_t$). We treat this as a practical stability and memory choice, not as evidence that zero momentum is generally optimal for RL.

This choice also avoids maintaining a persistent first-moment buffer for the Muon-updated actor weights during RL, reducing optimizer-state memory relative to momentum Muon. We did not include a controlled optimizer ablation in this report. A direct comparison against momentum Muon, AdamW, and SGD updates is left for future work.

\subsection{Monitoring and maintaining stability}
\label{sec:stability}

Reward and KL diagnostics describe the policy's optimization dynamics but do not reflect the content of generated rollouts. We monitor a small set of auxiliary rollout-level statistics during RL training to fill this gap. A subset of these statistics also act as reward gates, zeroing a rollout's task reward when its content is flagged as degenerate.

\paragraph{Streaming compressibility} Our primary canary is a sliding-window LZ77 compressibility metric computed per chunk on the raw token-ID bytes of each rollout. Compression uses \texttt{zlib} with a $2^{10} = 1024$-byte LZ77 window (\texttt{wbits=-10}), level-1 deflate, and \texttt{Z\_SYNC\_FLUSH} between chunks; the compressor is stateful, so each chunk's compression ratio reflects compressibility relative to recent history bounded by the LZ77 window rather than whole-sequence redundancy. Each rollout is divided into fixed-size chunks of $C$ tokens (with the final short chunk merged into its predecessor to avoid \texttt{Z\_SYNC\_FLUSH} overhead inflating short-tail ratios), and the per-chunk compression ratio
\begin{equation}
r_c = \frac{\text{compressed bytes}_c - \text{flush overhead}}{\text{raw token-ID bytes}_c}
\end{equation}
is computed for each chunk $c$.

A small $r_c$ indicates a chunk that compresses well against its preceding context, which is the signature of degenerate repetition or copying: the model has emitted a span of tokens already present in the LZ77 window. More generally, as is noted by \citep{lee2026nca}, an effective compression algorithm also serves as a computable upper bound on Kolmogorov Complexity \citep{li2019kolmogorov}, and both ends of the compressibility spectrum could arguably be filtered as either low information content or purely random. We choose LZ77 in particular over simpler n-gram or token counting methods because it takes into account sequence-level matching within the window, whereas language and domain-level n-gram biases can complicate simpler presence/frequency metrics. We flag a rollout if any chunk satisfies $r_c < \tau_{\text{repeat}}$, with a conservative $\tau_{\text{repeat}} = 0.05$ in production. Flagged rollouts have their task reward zeroed before advantage computation, so the policy receives no positive learning signal for producing degenerate text even when the verifier accepts the (technically correct) final answer at the end of a long repetitive trace. The per-chunk granularity allows reward zeroing on rollouts where degenerate spans appear at any position rather than attempting to rely on coarser, full response compressibility. 

\paragraph{Rare-token monitoring} As an independent signal, we track the fraction of tokens in each rollout whose token IDs fall in the top $X$\% of the tokenizer's ID range. This is a lightweight proxy for unusual or rarely used tokens in our tokenizer. In production monitoring we track several cutoffs, including 10\%, 5\%, 2\%, and 1\%, and use the top-10\% token-ID region for gibberish canaries. A rising rare-token fraction often precedes other failure indicators and is cheap to compute.

\paragraph{Operational use} The low-ratio repetition canary and rare-token-fraction statistics are computed per batch during RL training and visible alongside reward and KL in WandB. The repetition canary additionally runs as a reward-zeroing gate: rollouts that exceed the low-ratio threshold have their rewards zeroed before advantage computation, regardless of verifier outcome. Canary signals do not adjust learning rate or any other optimizer setting.

\paragraph{Length bias from signed KL-in-reward under pipeline RL}
Beyond rollout-level canaries, we also monitored response-length growth, which exposed an interaction between PipelineRL training and a sequence-level signed log-ratio reward penalty. In early stress tests combining the two, we observed runaway response-length growth: rollouts grew progressively longer over training without corresponding reward improvement. Our working explanation is specific to this estimator and aggregation choice. In pipeline RL, long completions can span multiple generator-policy snapshots: early tokens may be sampled from a stale generator policy $\pi_{\text{gen},c}$ that is $\Delta_c$ trainer updates behind the current actor $\pi_\theta$, while later tokens may be sampled from fresher snapshots with smaller $\Delta_c$.

The commonly used $K_1$-estimator log-ratio KL term is
\begin{equation}
l_t =
\log \pi_\theta(y_t \mid h_t)
-
\log \pi_{\text{gen},c(t)}(y_t \mid h_t).
\end{equation}
For tokens sampled from $\pi_{\text{gen},c(t)}$, this signed log-ratio is negative in expectation whenever the current policy differs from the generator policy:
\begin{equation}
\mathbb{E}_{y_t \sim \pi_{\text{gen},c(t)}}[l_t]
=
-
D_{\mathrm{KL}}\!\left(
\pi_{\text{gen},c(t)}(\cdot \mid h_t)
\,\|\, 
\pi_\theta(\cdot \mid h_t)
\right)
\le 0.
\end{equation}
For fresh tokens with small policy lag, $l_t \approx 0$. If these terms are aggregated into a sequence-level scalar,
\begin{equation}
S_{\text{seq}} = \sum_t l_t,
\end{equation}
and subtracted from reward as
\begin{equation}
A = r - \beta_{\mathrm{KL}} S_{\text{seq}},
\end{equation}
then stale off-policy tokens can create a positive reward offset. Longer completions can accumulate more negative signed log-ratio terms, and when the resulting sequence-level adjusted advantage is broadcast back to all tokens, stale-prefix terms can affect the learning signal assigned to later suffix tokens.

This produces a length-dependent bias through two interacting effects.

\textit{Stale-prefix contamination.}
Longer sequences can contain more stale prefix tokens contributing negative $l_t$, making $S_{\text{seq}}$ more negative. Since the signed log-ratio term enters as $-\beta_{\mathrm{KL}} S_{\text{seq}}$, the negative sequence sum acts as a positive reward offset, inflating the advantage for longer sequences independent of task quality.

\textit{Staleness-dependent penalty scale.}
The magnitude of the signed log-ratio can also depend on chunk staleness $\Delta_c$. Relatedly, Bartoldson derives a first-order EMA-reference approximation for asynchronous RL in which the log-ratio between the current policy and a $\Delta$-old inference policy can be interpreted as a surrogate for KL regularization against an EMA reference, under local linearity and first-order Taylor assumptions~\citep{bartoldson2026ema_kl}. This suggests that $\Delta$ can change the effective scale of a stale-policy log-ratio penalty. In our setting, we use this only as intuition for lag-dependent penalty strength; the length-bias mechanism itself follows from applying a signed off-policy log-ratio at sequence level and subtracting it from reward.

This mechanism should be distinguished from true KL regularization. A KL divergence is non-negative by construction, whereas the sampled signed log-ratio $l_t$ can be arbitrarily negative on individual samples and is negative in expectation under the generator distribution when $\pi_{\text{gen},c(t)} \ne \pi_\theta$. The severity depends on both absolute staleness and within-sequence policy heterogeneity. In-flight synchronization can create prefix--suffix heterogeneity by allowing one completion to span multiple generator snapshots; holding the generator fixed for the entire completion removes this specific coupling, but does not remove the off-policy signed-log-ratio length offset if the fixed generator is stale relative to the trainer.

\paragraph{Possible mitigations}
Two practical mitigations target the specific stale-prefix coupling described above. \textit{Chunk-local signed-log-ratio isolation} aggregates the signed log-ratio within each chunk rather than across the full sequence, so stale-prefix terms do not directly contaminate the advantage assigned to fresher suffix chunks:
\begin{equation}
\mathcal{A}_c
=
A_{\text{reward}}
-
\beta_{\mathrm{KL}} S_c,
\qquad
S_c = \sum_{t \in c} l_t .
\end{equation}
This localizes the bias but does not by itself turn the off-policy signed log-ratio into a true KL penalty.

\textit{Staleness rescaling} is an additional heuristic: divide the chunk term by an empirical staleness scale $g(\Delta_c)$, with $g(\Delta_c) > 0$, to reduce variation in effective penalty strength across chunks generated at different lags:
\begin{equation}
\mathcal{A}_c
=
A_{\text{reward}}
-
\beta_{\mathrm{KL}}
\cdot
\frac{1}{g(\Delta_c)}
\sum_{t \in c} l_t .
\end{equation}
A simple first-order choice is $g(\Delta_c)=\max(1,\Delta_c)$, motivated by the local-linear lag dependence in Bartoldson's EMA approximation~\citep{bartoldson2026ema_kl}, but the correct scale is implementation- and dynamics-dependent.

For ZAYA1-8B we did not implement either mitigation in production. Instead, we removed KL-in-reward entirely and rely on the DPPO Binary-TV trust region (Section~\ref{sec:rl-algo}) for trust-region enforcement. This was sufficient for the training-stability properties we required and avoided tracking chunk boundaries and per-chunk generator staleness. We document the mechanism here because it may arise in asynchronous or pipeline RL systems that combine stale or mixed-policy rollouts, a signed $K_1$-estimator log-ratio in the reward, sequence-level aggregation, and broadcast of the resulting adjusted advantage.

\section{Results}
\label{sec:results}

\begin{figure*}
    \centering
    \includegraphics[width=1\linewidth]{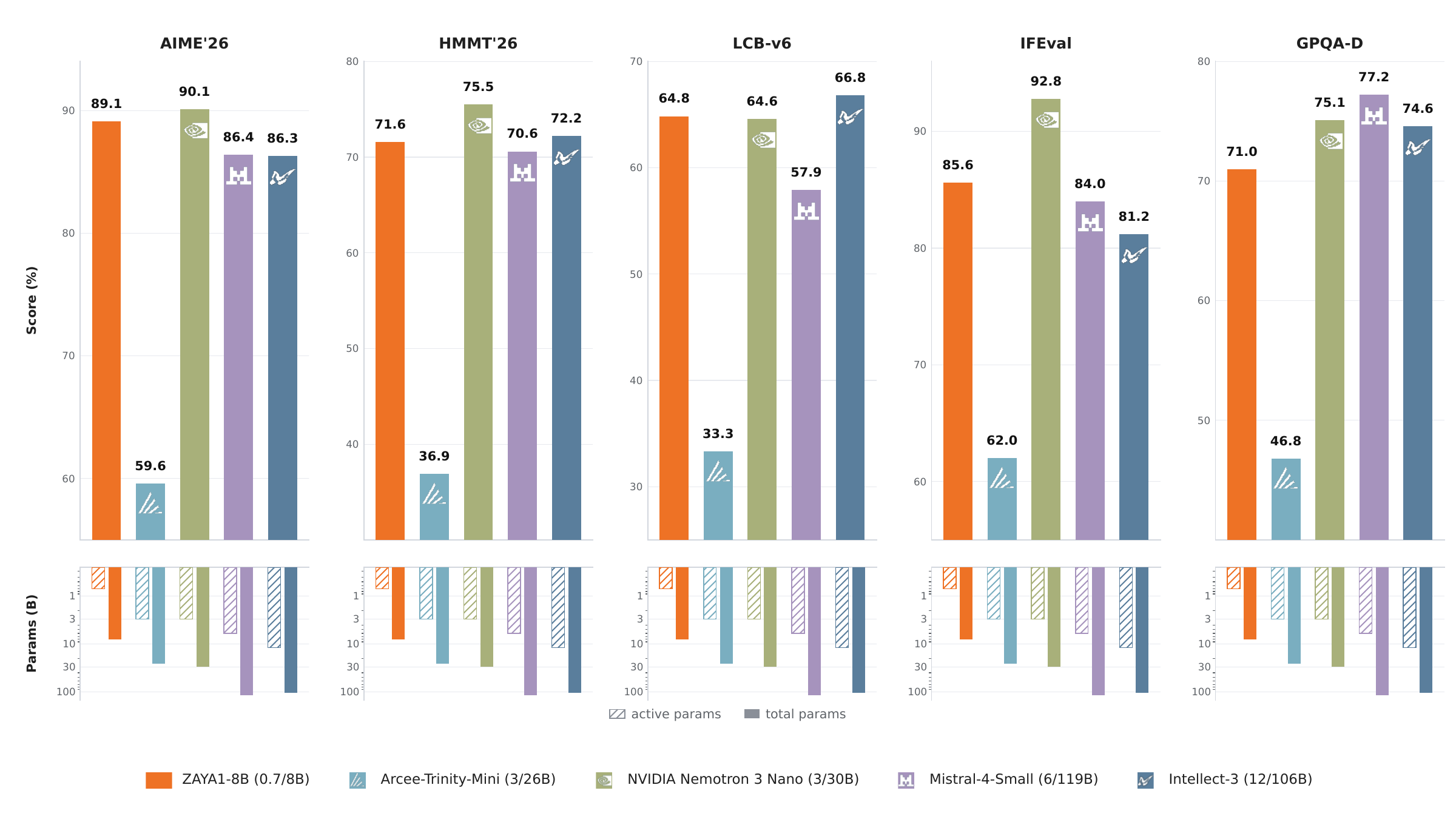}
    \caption{Comparison of ZAYA1-8B performance against open-weight reasoning models on various evaluations. The under-bar plots model sizes in active and total parameters on a log scale to give a sense of the scale of the various models.}
    \label{western_OS_models}
\end{figure*}

Results are organized into three tables. Table~\ref{tab:results-inclass} compares ZAYA1-8B against open-weight reasoning models at comparable scale. Table~\ref{tab:results-scaling} extends to open-weight models in the 26B--119B total-parameter range. Table~\ref{tab:results-ttc} reports test-time compute comparisons against open-weight models in the 235B--671B range plus Gemini-2.5 Pro and GPT-5-High.

\subsection{Evaluation Protocol}
\label{sec:eval-protocol}

Unless otherwise noted, ZAYA1-8B results are measured with the Zyphra evaluation harness. In-class comparator models are run in the same harness when feasible, using each model's recommended sampling settings from its model card. For ZAYA1-8B single-rollout reasoning evaluations, we use temperature 1.0, top-p 0.95, top-k -1, and benchmark-specific maximum generation lengths. For thinking-mode Qwen comparators, we mirror the recommended thinking-mode settings from the corresponding model card. Results reported from external release materials are marked with $^\dagger$. TTC evaluations use the checkpoint immediately following the math+code+TTC RL stage and before the final behavioral-RL polishing stage; the latter targets chat style, instruction following, and preference behavior rather than additional math/code/TTC capability.

For pass-rate evaluations in the Zyphra harness, we report averages over multiple samples per problem. Math benchmarks, including AIME, HMMT, IMO-AnswerBench, and APEX-shortlist, are reported as avg@64. Code benchmarks, including LiveCodeBench-family tasks, are reported as avg@16. GPQA-Diamond and $\tau^2$ are reported as avg@16 unless otherwise noted. MMLU-Pro, BFCL-v4, HLE, IFEval, IFBench, EQBench, and Creative Writing are reported as mean@1 or as the benchmark's standard single-run score. We use avg@k to mean the mean correctness over k independently sampled completions, estimating single-sample pass rate under the stated sampler; it is not best-of-k/pass@k unless explicitly stated. Markovian RSA results use the TTC protocol in Section~\ref{sec:ttc}; its token counts are total newly generated decode tokens and exclude prompt/prefill tokens. Results from external release materials may use different sampling and reporting protocols and are marked with $^\dagger$.

\subsection{Main Results: In-Class Comparison}
\label{sec:results-inclass}

\begin{table*}[htbp]
\centering
\small
\begin{tabular}{llcccc}
\toprule
\multicolumn{2}{l}{Active / Total} & 0.7B / 8.0B & 4.0B / 4.0B & 4.0B / 4.0B & 4.0B / $^*$8.0B \\
\midrule
Category & Benchmark & ZAYA1-8B & Qwen3-4B-Thinking-2507 & Qwen3.5-4B & Gemma-4-E4B-it \\
\midrule
\multirow{4}{*}{Math}
 & AIME'26                       & 89.1 & 79.0 & 84.5 & 50.3 \\
 & HMMT'26 Feb.                  & 71.6 & 53.6 & 63.6   & 32.1 \\
 & IMO-AnswerBench               & 59.3 & 51.6 & 48.7   & 27.3 \\
 & APEX-shortlist                & 32.2 & 17.1 & 21.35   & 6.1   \\
\midrule
\multirow{1}{*}{Code}
 & LiveCodeBench-v6              & 64.8 & 54.9 & 55.8$^\dagger$   & 54.2 \\
\midrule
\multirow{2}{*}{Knowledge}
 & GPQA-Diamond                  & 71.0 & 66.1 & 76.2   & 57.4 \\
 & MMLU-Pro                      & 74.2 & 74.3 & 79.7   & 70.2 \\
\midrule
\multirow{2}{*}{Instruction}
 & IFEval                        & 85.6 & 86.8 & 89.8 & 88.5   \\
 & IFBench                       & 52.6 & 52.9 & 59.2 & 42.7   \\
\midrule
\multirow{2}{*}{Style \& chat}
 & EQBench                       & 73.0 & 79.6 & 79.5 & 80.2   \\
 & Creative Writing v3           & 63.0 & 58.6 & 72.9 & 83.8   \\
\midrule
\multirow{2}{*}{Agentic}
 & BFCL-v4                       & 40.5 & 49.7 & 45.2 & 31.7 \\
 & $\tau^2$                      & 36.3 & 52.9 & 82.1 & 37.7 \\
\bottomrule
\end{tabular}

\vspace{4pt}
{\raggedright \footnotesize $^*$Gemma4 includes 4B additional embedding parameters as a part of its total.\par}
{\raggedright \footnotesize $^\dagger$Qwen3.5-4B LiveCodeBench-v6 scores taken from release materials.\par}
\caption{In-class comparison against models of comparable sizes. ZAYA1-8B used the following sampling settings: $T{=}1.0$, top-$p{=}0.95$, top-$k$ disabled for math, knowledge, and instruction; $T{=}0.6$, top-$p{=}0.95$, top-$k{=}20$ for code, agentic, and style. We used the recommended sampling settings in the model cards for the other models in this table. EQBench and Creative Writing v3 use the official judge, \texttt{anthropic/claude-3.7-sonnet}.} 
\label{tab:results-inclass}
\end{table*}
Table~\ref{tab:results-inclass} compares ZAYA1-8B against Qwen3-4B-Thinking-2507, Qwen3.5-4B, and Gemma-4-E4B-it.
\subsection{Scaling Comparison: Larger Open-Weight Models}
\label{sec:results-scaling}

Table~\ref{tab:results-scaling} compares ZAYA1-8B against larger open-weight reasoning models: Arcee-Trinity-Mini, Nemotron-3-Nano, OLMo-3.1-32B-Think, Qwen3-Next-80B-A3B-Think, Intellect-3, and Mistral-Small-4-119B-2603.

\begin{table*}[h]
\centering
\small
\begin{tabular}{lrrcccccc}
\toprule
Model & Active & Total & AIME'26 & HMMT'26 Feb. & LCB-v6* & IFEval & GPQA-D & MMLU-Pro \\
\midrule
ZAYA1-8B                  & 0.7B & 8B    & 89.1 & 71.6 & 64.8 & 85.6 & 71.0 & 74.2 \\
Arcee-Trinity-Mini        & 3B   & 26B   & 59.6 & 36.9 & 33.3 & 62.0 & 46.8 & 70.6 \\
Nemotron-3-Nano-30B-A3B   & 3B   & 30B   & 90.1 & 75.5 & 64.6 & 92.8 & 75.1 & 78.9 \\
OLMo-3.1-32B-Think        & 32B  & 32B   & 78.9 & 50.6 & 58.3 & 93.2 & 59.6 & 75.8 \\
Qwen3-Next-80B-A3B-Think  & 3B   & 80B   & 90.2 & 79.3 & 67.8 & 88.5 & 76.7 & 82.6 \\
Intellect-3               & 12B  & 106B  & 86.3 & 72.3 & 66.8 & 81.2 & 74.6 & 82.3 \\
Mistral-Small-4-119B      & 6B   & 119B  & 86.4 & 70.6 & 57.9 & 84.0 & 77.2 & 81.6 \\
\bottomrule
\end{tabular}
\caption{Scaling comparison against larger open-weight reasoning models, ordered by total parameter count. All numbers are run on the Zyphra evaluation harness.}
\label{tab:results-scaling}

\vspace{4pt}
{\raggedright \footnotesize $^*$LCB-v6 denotes the 2025-02--2025-05 LiveCodeBench-v6 split.\par}
\end{table*}
\subsection{Test-Time Compute Scaling}
\label{sec:results-ttc}

Table~\ref{tab:results-ttc} compares ZAYA1-8B with Markovian RSA test-time compute against substantially larger reasoning models. With the headline Markovian RSA configuration ($\beta$ = 40K, $\tau$ = 4K, $T$ = 2, $N$ = 16, $C$ = 4), ZAYA1-8B reaches 91.9 on AIME'25 and 89.6 on HMMT'25 Feb.

\subsection{Effect of Post-Training}
\label{sec:results-posttraining}

To quantify the effect of post-training, we compare the 131K SFT checkpoint against the final ZAYA1-8B checkpoint using the same evaluation harness and sampling settings in Table~\ref{tab:sft-to-final}. This comparison measures the aggregate effect of the RL cascade rather than isolating the contribution of each individual stage. We do not report per-stage ablations in this release.

\begin{table*}[h]
\centering
\small
\begin{tabular}{lccc}
\toprule
Benchmark & SFT checkpoint & Final ZAYA1-8B & $\Delta$ \\
\midrule
AIME'26              & 68.30 & 89.10 & +20.80 \\
HMMT'26 Feb.        & 39.20 & 71.60 & +32.40 \\
LiveCodeBench-v6     & 54.80 & 64.84 &  +10.04 \\
GPQA-Diamond         & 59.30 & 71.00 & +11.70 \\
MMLU-Pro             & 70.10 & 74.20 &  +4.10 \\
IFEval               & 66.60 & 85.58 & +18.98 \\
IFBench              & 30.20 & 52.56 & +22.36 \\
EQBench              & 57.80 & 72.95 & +15.15 \\
Creative Writing v3  & 46.72 & 62.97 & +16.25 \\
BFCL-v4              & 33.41 & 40.50 & +7.09  \\
$\tau^2$             & 32.88 & 36.30 & +3.42  \\
\bottomrule
\end{tabular}
\caption{Aggregate effect of post-training. SFT and final ZAYA1-8B checkpoints are evaluated with the same harness and benchmark-specific sampling settings. This table reports the aggregate effect of the post-training recipe; it is not a per-stage ablation.}
\label{tab:sft-to-final}
\end{table*}

\section{Test-Time Compute}
\label{sec:ttc}
Test-time compute (TTC) scaling --- increasing inference compute per problem to improve answer quality --- has become an important axis of capability scaling for reasoning models, alongside model scale and training compute. Two recent lines of work motivate the design space considered here. \citep{rsa} introduce Recursive Self-Aggregation (RSA), a TTC scheme that maintains a population of candidate reasoning chains and refines them through repeated aggregation: at each iteration, the model is shown a random subset of candidates and produces an improved candidate, which seeds the next iteration's population. Empirically, RSA allows smaller open-weight models to approach the performance of larger reasoning models when given sufficient inference compute. \citep{markovian-thinker} introduce Markovian Thinker, a reformulation of the RL thinking environment in which the policy reasons in fixed-size chunks with bounded carryover state between chunks, decoupling thinking length from context size. Their key observation is that long-context reasoning can be factorized in a Markovian way: with sufficient training, a model can sometimes carry forward only the information needed in a bounded textual state and continue reasoning indefinitely.

We introduce \emph{Markovian RSA}, a TTC method that combines RSA's recursive candidate aggregation with the bounded-workspace principle of Markovian Thinker. We integrate it into ZAYA1-8B's training pipeline so the model is trained to use the same workflow at inference. The method has three components: an algorithm that includes both RSA and Markovian-Thinker for chunked reasoning as special cases (Section~\ref{sec:ttc-algorithm}), a training-time integration that supplies verifier-free aggregation examples for SFT and verifiable aggregation prompts for RL (Section~\ref{sec:ttc-training}), and an inference-time scaling profile with bounded per-iteration aggregation context, capped attention costs, and predictable throughput (Section~\ref{sec:ttc-inference}).

\subsection{Markovian RSA}
\label{sec:ttc-algorithm}
\paragraph{Algorithm} Given a problem $q$ and a base policy $\pi$, Markovian RSA proceeds over $T$ aggregation rounds, indexed $t = 0, 1, \ldots, T$. Each round maintains a population of $N$ candidate reasoning traces. At round $t=0$, the model generates $N$ independent rollouts directly from $q$, each with a per-rollout thinking budget $\beta$. Each rollout's reasoning trace is then reduced to its last $\tau$ tokens, which we call the \emph{tail}. We write $\text{tail}_\tau(y)$ for the operation that returns the final $\tau$ tokens of reasoning trace $y$, with $\tau \leq \beta$.

For rounds $t \geq 1$, the algorithm operates on tails from the previous population. To generate each new candidate, it samples $C \leq N$ tails uniformly at random, concatenates them into an aggregation prompt, and asks the model to reason over the candidate solutions and produce a single improved solution. The model generates a new reasoning trace under the same per-rollout budget $\beta$. The trace is again reduced to its final $\tau$ tokens, and the resulting tail enters the population for round $t$. This process repeats until round $T$, after which the final answer is extracted from the final round's outputs using the standard answer-extraction procedure. The aggregation prompt simply asks the model to consider the candidates and produce the best solution; it does not require specialized parsing or verifier feedback.

Both Markovian RSA and full-chain RSA bound per-rollout generation cost: $\beta$ caps the number of tokens any single candidate generates. The difference is what gets passed forward. Full-chain RSA passes the full reasoning chain\footnote{One can add a summarization step to full-chain RSA to keep aggregation prompts short. We focus on fixed-tail forwarding because it gives a simple, bounded aggregation context without requiring an additional summarization model or parsing step.}, so the aggregation prompt at round $t \geq 1$ contains $C$ chains, each with length up to $\beta$. Markovian RSA passes only the final $\tau$ tokens of each chain, with $\tau \leq \beta$ chosen independently. This decouples per-rollout thinking depth from aggregation-context size: $\beta$ controls how long each candidate may reason, while $\tau$ controls how much of that reasoning is carried into the next round. Setting $\tau \ll \beta$ allows larger per-rollout thinking budgets while keeping aggregation prompts small. As a result, decode-attention cost, prefill-attention cost, and KV-cache footprint are bounded by configuration constants rather than by reasoning length.

\paragraph{Default configuration} For ZAYA1-8B, we use $(N, C, T) = (16, 4, 2)$ with $\beta$ set per workload and $\tau$ chosen as a fraction of $\beta$ (typically $\tau \leq \beta/2$). Both $\beta$ and $\tau$ can be tuned per deployment to trade off per-round thinking depth against total inference budget.

\paragraph{Inference profile}
Markovian RSA changes the inference workload from a single long, position-growing decode into a sequence of bounded-context batched decoding stages. At round $0$, the model generates $N$ independent candidates from the original problem, so decode runs at batch size $N$ rather than batch size $1$. At each later aggregation round, the model again generates $N$ candidates, but each candidate conditions only on the problem and $C$ carried-forward tails of length at most $\tau$. Thus the aggregation prefill length is bounded by
\begin{equation}
    L_{\mathrm{prefill}} \leq |q| + C\tau + O(1),
\end{equation}
and the per-candidate decode length is bounded by $\beta$, independent of the total amount of reasoning generated across all rounds. This gives a stable serving profile: prefill is short and predictable at every stage, decode uses high-throughput batched generation, and no stage attends over the full reasoning history.

This profile differs from both single-rollout long-CoT and full-chain RSA. A single long rollout has batch size $1$ and a decode position that grows with the full reasoning length. Full-chain RSA supports batched candidate generation, but its aggregation prefill grows with $C\beta$ because it forwards full reasoning chains. Markovian RSA keeps the batched candidate-generation structure of RSA while replacing full-chain forwarding with bounded tail forwarding, so increasing $\beta$ increases per-candidate thinking depth without increasing aggregation-context length.

\begin{table*}[h]
\centering
\small
\setlength{\tabcolsep}{4pt}
\begin{tabularx}{\linewidth}{@{}
  >{\RaggedRight\arraybackslash}p{0.23\linewidth}
  >{\RaggedRight\arraybackslash}X
  >{\RaggedRight\arraybackslash}X
  >{\RaggedRight\arraybackslash}p{0.19\linewidth}
@{}}
\toprule
Method & Decode profile & Aggregation/context state & Special case \\
\midrule
Single long-CoT
& BS=1, position grows with total reasoning
& Full prior trace
& -- \\

Parallel sampling / $N$ responses
& BS=$N$, one round
& No aggregation
& Markovian RSA with $T=0$; becomes Best-of-$N$ only with an external selector or verifier \\

Delethink continuation
& BS=$N$, chunked rounds
& One bounded tail
& Markovian RSA with $C=1$ \\

Full-chain RSA
& BS=$N$, aggregation rounds
& $C$ full chains, length up to $C\beta$
& Markovian RSA with $\tau=\beta$ \\

Markovian RSA
& BS=$N$, aggregation rounds
& $C$ tails, length up to $C\tau$
& General case \\
\bottomrule
\end{tabularx}
\caption{Inference-profile view of Markovian RSA. Markovian RSA preserves the batched candidate-generation structure of RSA while bounding the state forwarded between rounds. Setting $T=0$ gives $N$ independent responses with no aggregation; this becomes Best-of-$N$ only if an external selector, verifier, or answer-selection rule is applied. Setting $C=1$ gives the Delethink bounded-continuation regime, and setting $\tau=\beta$ recovers full-chain RSA.}
\label{tab:markovian-rsa-profile}
\end{table*}

\paragraph{Special cases}
Markovian RSA contains several common TTC regimes as special cases:

\begin{itemize}[leftmargin=*,itemsep=2pt,topsep=2pt]
    \item \textbf{Parallel sampling / $N$ responses.} Setting $T=0$ removes aggregation and produces $N$ independent responses. If a verifier, answer-selection rule, or external scoring model is applied to these responses, this reduces to a Best-of-$N$ evaluation; otherwise it is simply parallel sampling.
    \item \textbf{Full-chain RSA.} Setting $\tau=\beta$ forwards each full reasoning chain between rounds, recovering RSA. In this limit, aggregation prefill grows with the full reasoning budget.
    \item \textbf{Delethink bounded continuation.} Setting $C=1$ removes cross-candidate aggregation while retaining bounded carryover. Each candidate continues from its own tail, giving a parallel version of Markovian/Delethink chunked reasoning. This isolates the effect of bounded continuation from the additional effect of cross-candidate aggregation.
\end{itemize}

\paragraph{Comparison with PaCoRe}
\citep{pacore} introduced PaCoRe, a related multi-round parallel-reasoning scheme that also bounds per-round aggregation context. PaCoRe compacts each trajectory by extracting its final-answer or conclusion section and passing this extracted message forward between rounds. Markovian RSA instead passes the final $\tau$ tokens of the reasoning trace itself as the carry-forward state, regardless of whether the trajectory reached a final conclusion. The two methods share the same goal of bounding aggregation context across rounds and differ in compaction mechanism: PaCoRe uses model-structured final-answer extraction, while Markovian RSA uses a fixed-size suffix of generated reasoning.

In practice, we also evaluate a PaCoRe hybrid compaction variant: when a candidate reaches a post-think answer section, we pass that compact answer forward; otherwise, we fall back to passing the partial reasoning chain. This hybrid keeps the compact-message advantage of PaCoRe when candidates finish, while avoiding the need to set $\beta$ large enough for every branch to reach a final answer.

\begin{figure*}[h]
    \centering
    \includegraphics[width=0.95\linewidth]{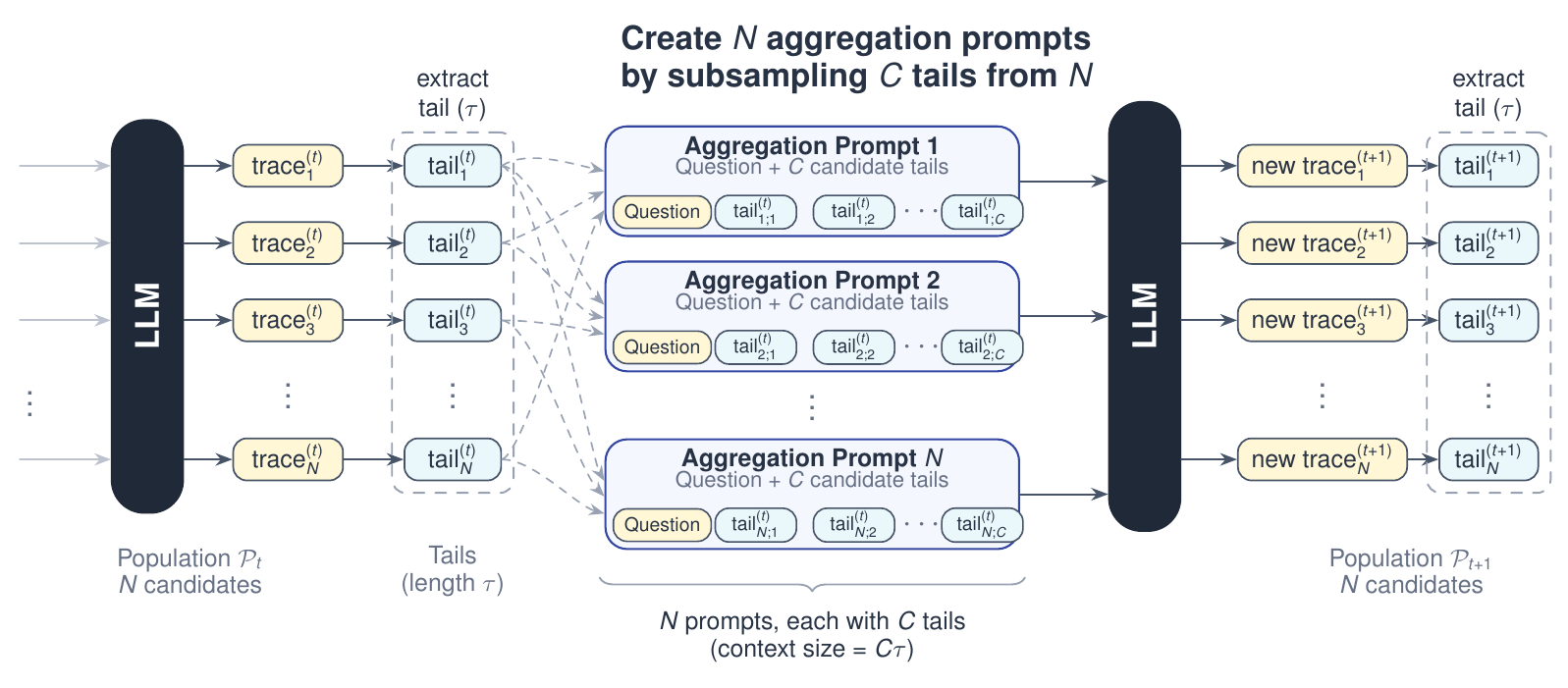}
    \caption{One round of Markovian RSA. From a population of $N$ candidate reasoning traces (left), we extract the final $\tau$ tokens of each trace as its tail. To produce each new candidate for the next round, we sample $C$ tails uniformly at random and present them to the model as candidate solutions in an aggregation prompt. The model produces a new reasoning trace, whose tail joins the next round's population. Aggregation context size and per-round attention cost depend only on $C$ and $\tau$, and are independent of the per-rollout thinking budget $\beta$.}
    \label{fig:ttc-algorithm}
\end{figure*}

\subsection{Training-Time Integration}
\label{sec:ttc-training}

A TTC method may be more effective when the model is trained on the workflow it uses at inference. Markovian RSA's aggregation prompt presents the model with a problem and several candidate reasoning tails, then asks it to produce a single improved solution. This behavior is rare in standard reasoning-model training data, where each example typically consists of one problem and one solution. To train ZAYA1-8B for Markovian RSA scaling, we construct aggregation-based examples from existing expert-model reasoning data and include them in SFT and RL.

\paragraph{SFT data construction from expert rollouts} Many open-source reasoning datasets used during midtraining and SFT include multiple expert-model rollouts per problem, often with $n=8$ rollouts (e.g., \texttt{OpenMathReasoning}, \texttt{rStar-Coder}, internal reasoning gym and enigmata data). For each problem $q$ with rollouts $\{y_1, \ldots, y_n\}$ from a teacher model, we construct a round-0-to-round-1 aggregation example as follows: sample $C$ rollouts from the $n$ available; extract their tails $\{\text{tail}_\tau(y_{i_1}), \ldots, \text{tail}_\tau(y_{i_C})\}$; form an aggregation prompt containing $q$ and the $C$ tails; and condition the teacher to produce a new aggregated rollout under the same prompt. The resulting aggregated rollout, including its reasoning trace and final answer, becomes the SFT target.

This construction has two practical advantages. It is offline and reuses existing rollout pools: no new expert-model inference is needed for each round-0 sample beyond the aggregation step. It also does not require a verifier: the teacher's aggregated rollout is used as the target regardless of whether the underlying answer is verifiable. This makes the technique applicable to puzzle, code, and reasoning domains where the post-think content is itself the answer and where final-answer-only aggregation strategies such as PaCoRe's message compaction are not directly applicable.

\paragraph{RL stage integration} During RL, Markovian RSA examples are folded into the standard prompt distribution and treated like other RL prompts. Two variants are used during the math+code+TTC stage (Section~\ref{sec:rl-mathcodettc}):

\begin{itemize}[leftmargin=*,itemsep=2pt,topsep=2pt]
    \item \textbf{Expert-aggregation.} Round-1 prompts are constructed from expert-model rollouts as described above. The policy generates an aggregated rollout and is rewarded against the verifiable target.
    \item \textbf{Self-aggregation.} For prompts where rollouts from the current SFT checkpoint or a prior-stage RL checkpoint are available, round-1 prompts are constructed from those self-rollouts. The policy aggregates over its own reasoning traces, or over traces from its predecessor.
\end{itemize}

In both variants, the aggregation example is a standard RL prompt: the policy generates a single rollout, and verifiable reward is applied to its final answer. No special multi-round RL machinery is required; the round structure is encoded in the prompt construction rather than in the gradient update. We currently train on round-1 self-aggregation. Round-2-and-beyond self-aggregation, where the policy aggregates rollouts from a prior-stage version of itself in an online buffer, is a natural extension left for future work.

\paragraph{Domain coverage} Aggregation-based training data is included for math, code, reasoning gym, and enigmata puzzle problems. Directly aggregating over reasoning tails is useful in domains where the post-think content is the answer rather than a separate boxed result. This allows the same approach to apply across domains regardless of answer format.

\subsection{Inference-Time Scaling}
\label{sec:ttc-inference}

\begin{table*}[h]
\centering
\small
\begin{tabular}{lrrccc}
\toprule
Model & Active & Total & AIME'25 & HMMT'25 Feb. & LCB-v6* \\
\midrule
ZAYA1-8B (single rollout)               & 0.7B & 8.0B  & 88.3 & 82.7 & 65.0 \\
ZAYA1-8B + Markovian RSA (40K/4K)       & 0.7B & 8.0B  & 91.9 & 89.6 & 69.2$^\ddagger$  \\
\midrule
DeepSeek-R1-0528$^\dagger$              & 37B  & 671B  & 87.5 & 79.4 & 68.7 \\
Qwen3-235B-A22B-Thinking-2507$^\dagger$ & 22B  & 235B  & 92.3 & 83.9 & 74.1 \\
Gemini-2.5 Pro$^\dagger$                & --   & --    & 88.0 & 82.5   & 72.5   \\
DeepSeek-V3.2$^\dagger$                 & 37B  & 671B  & 93.1 & 92.5   & --   \\
GPT-5-High$^\dagger$                    & --   & --    & 94.6 & 88.3   & --   \\
\bottomrule
\end{tabular}

\vspace{4pt}
{\raggedright \footnotesize $^*$LCB-v6 denotes the 2025-02--2025-05 LiveCodeBench-v6 split.\par}

\vspace{4pt}
{\raggedright \footnotesize $^\ddagger$ For LCB-v6 on the same pre-behavioral checkpoint after math+code+TTC RL, Markovian RSA improves ZAYA1-8B from 65\% single-rollout to 69.2\%, while our PaCoRe hybrid compaction variant reaches 71.1\%. This variant is not an exact implementation of PaCoRe: when a candidate reaches a post-think answer section, we pass that compact answer forward; when it does not, we fall back to passing the partial reasoning chain rather than dropping the candidate. Since the model was trained with both RSA and PaCoRe-type aggregation examples, we do not attribute the gap to training exposure alone.\par}

\caption{ZAYA1-8B single-rollout and TTC numbers in this table are evaluated on the pre-behavioral checkpoint after math+code+TTC RL and before the final lightweight behavioral-RL polishing stage, using the Zyphra evaluation harness. The final behavioral stage targets chat style, instruction following, and preference behavior rather than math/code/TTC capability. ZAYA1-8B + Markovian RSA uses the 40K/4K configuration from Section~\ref{sec:ttc-inference}. Numbers for comparator models marked $^\dagger$ are taken from external sources. DeepSeek-R1-0528, Qwen3-235B-A22B-Thinking-2507, and Gemini-2.5 Pro are from the Qwen3-235B-A22B-Thinking-2507 model card~\citep{qwen3thinking2507card}. DeepSeek-V3.2 is from the DeepSeek-V3.2 technical report~\citep{deepseekai2025deepseekv32}; the GPT-5-High row is reproduced from the comparison table in that report rather than from an OpenAI release table.}
\label{tab:results-ttc}
\end{table*}


We evaluate Markovian RSA's inference-time scaling along two axes --- per-rollout reasoning budget $\beta$ and tail size $\tau$ --- with the configuration described in Section~\ref{sec:ttc-algorithm} $(N, C, T) = (16, 4, 2)$. The sampling settings are reported in Table~\ref{tab:markovian-rsa-scaling}. We also compare against full-chain RSA \citep{rsa}, recovered as the $\tau = \beta$ limit of our algorithm. The final scores reported in this section are mean correctness over the final-round candidate outputs, not best-of-$N$ unless explicitly stated.
\paragraph{Headline result} With Markovian RSA at $(\beta, \tau, T, N, C) = (40\mathrm{K}, 4\mathrm{K}, 2, 16, 4)$, ZAYA1-8B reaches 91.9\% on AIME'25 and 89.6\% on HMMT'25 Feb. These runs use temperature 1.0, top-p 1.0, and a 40K-token final-response budget. The result holds while carrying forward only a 4K-token tail between aggregation rounds, approximately one-tenth of the per-rollout reasoning budget.

\paragraph{Configuration sweep} Table~\ref{tab:markovian-rsa-scaling} reports accuracy across four Markovian RSA configurations, alongside the $C=1$ Markovian Thinker baseline described in Section~\ref{sec:ttc-algorithm}. At $T=2, N=16$, and $C=4$, increasing the per-rollout reasoning budget $\beta$ from 8K to 16K to 40K improves both benchmarks at fixed tail size: AIME'25 advances from 86.5\% to 88.8\% to 91.9\%, and HMMT'25 advances from 80.8\% to 87.1\% to 89.6\%. HMMT'25 is especially responsive to longer per-rollout reasoning, with a 6.3-point gain from $\beta=8\text{K}$ to $\beta=16\text{K}$.

\begin{table*}[h]
\centering
\small
\begin{tabular}{lccccccc}
\toprule
Configuration & $T$ & $N$ & $C$ & $\beta$ (CoT) & $\tau$ (Tail) & AIME'25 & HMMT'25 \\
\midrule
Markovian baseline & 4 & 16 & 1 & 8K  & 4K & 82.1 & 75.0 \\
\midrule
Markovian RSA & 2 & 16 & 4 & 8K  & 4K & 86.5 & 80.8 \\
Markovian RSA & 2 & 16 & 4 & 16K & 4K & 88.8 & 87.1 \\
Markovian RSA & 2 & 16 & 4 & 40K & 4K & \textbf{91.9} & \textbf{89.6} \\
Markovian RSA & 2 & 16 & 4 & 40K & 8K & 90.8 & 89.2 \\
\bottomrule
\end{tabular}
\caption{Markovian RSA configuration sweep on AIME'25 and HMMT'25. $T$ is the number of aggregation rounds, $N$ is the population size, $C$ is the number of candidates sampled per aggregation prompt, $\beta$ is the per-rollout reasoning budget, and $\tau$ is the carry-forward tail size. Markovian RSA rows use temperature 1.0 and top-p 1.0 with a 40K-token final-response budget. The Markovian baseline row uses $C=1$, removing cross-candidate aggregation while retaining $N=16$ independent chunked rollouts.}
\label{tab:markovian-rsa-scaling}
\end{table*}

\paragraph{Evaluation scope}
The algorithmic definition, training construction, and serving profile above describe the Markovian RSA method. The remainder of this subsection reports empirical TTC scaling results, measuring how accuracy changes with the per-rollout reasoning budget $\beta$, carried-forward tail length $\tau$, aggregation depth $T$, and aggregation method under a fixed population size.

\paragraph{Generated-token accounting}
We report realized total generated decode tokens separately from per-worker trajectory lengths. Markovian RSA generates multiple candidates in parallel at each non-final stage, so a per-worker or per-stage average length is not the total decode cost of the method. For a problem $q$, let $g_{s,j}(q)$ be the number of newly generated tokens from worker $j$ in generation stage $s$, and let $n_s$ be the number of workers in that stage. The realized generated-token cost is
\begin{equation}\label{eqn:ttc-token-count}
D(q)
=
\sum_s \sum_{j=1}^{n_s} g_{s,j}(q)
=
\sum_s n_s \bar{g}_s(q),
\end{equation}
where $\bar{g}_s(q)$ is the average generated length per worker in stage $s$. This count includes newly generated candidate and aggregation tokens across all workers, but excludes the original problem prompt, aggregation-prompt prefill, and copied carry-forward tails.

Under this accounting, with the final response budget of 40K, the reported AIME'25/HMMT'25 Markovian RSA evaluations use approximately 440K generated decode tokens per problem for the 16K/4K configuration and approximately 740K generated decode tokens per problem for the 40K/4K configuration. These are realized averages from the evaluation runs, not worst-case caps; they depend on early stopping, benchmark, sampling settings, and implementation details. We include them to avoid conflating per-worker trajectory length with total TTC cost.

\begin{figure}[h]
    \centering
    \begin{subfigure}{0.24\textwidth}
         \centering
         \includegraphics[width=\linewidth]{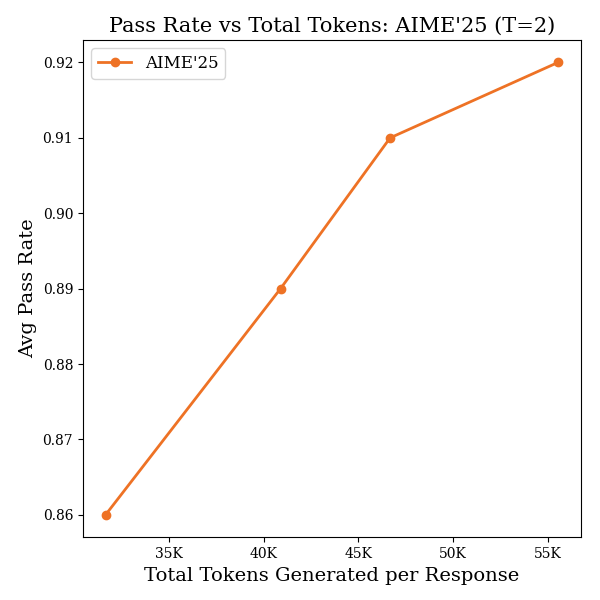}
     \end{subfigure}
     \begin{subfigure}{0.24\textwidth}
         \centering
         \includegraphics[width=\linewidth]{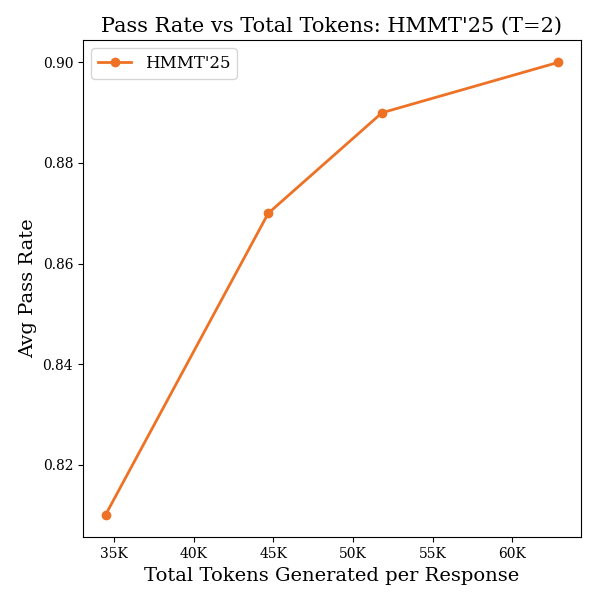}
     \end{subfigure}
     \caption{Accuracy vs.\ realized total newly generated decode tokens per problem for Markovian RSA configurations on AIME'25 (left) and HMMT'25 (right). The token axis excludes the original problem prompt, aggregation-prompt prefill, and copied carry-forward tails, and corresponds to $\sum_s \bar{g}_s(q)$, using notations from \eqref{eqn:ttc-token-count}, averaged over problems. Curves correspond to the configurations in Table~\ref{tab:markovian-rsa-scaling}: $\beta=8\text{K}, \tau=4\text{K}$; $\beta=16\text{K}, \tau=4\text{K}$; $\beta=40\text{K}, \tau=8\text{K}$; and $\beta=40\text{K}, \tau=4\text{K}$. All curves use $N=16$, $C=4$, $T=2$, temperature 1.0, and top-p 1.0. Allocating budget to longer per-rollout reasoning yields more accuracy per generated token than allocating to longer tail carryover in this sweep.}
     
    \label{fig:ttc-headline-scaling}
\end{figure}

\paragraph{Tail size and iteration depth}
Increasing the tail size $\tau$ from 4K to 8K at fixed $\beta=40\text{K}$ does not improve accuracy on AIME'25 or HMMT'25: the 4K-tail configuration reaches 91.9\%/89.6\%, while the 8K-tail configuration reaches 90.8\%/89.2\%. Because these scores average over multiple generated candidates per problem, the comparison is less sensitive to a single unlucky rollout than a one-sample evaluation. We treat this as empirical evidence that, for these configurations and benchmarks, aggregation is not limited by retaining more than a 4K reasoning tail.

\begin{figure}[h]
    \centering
    \includegraphics[width=\linewidth]{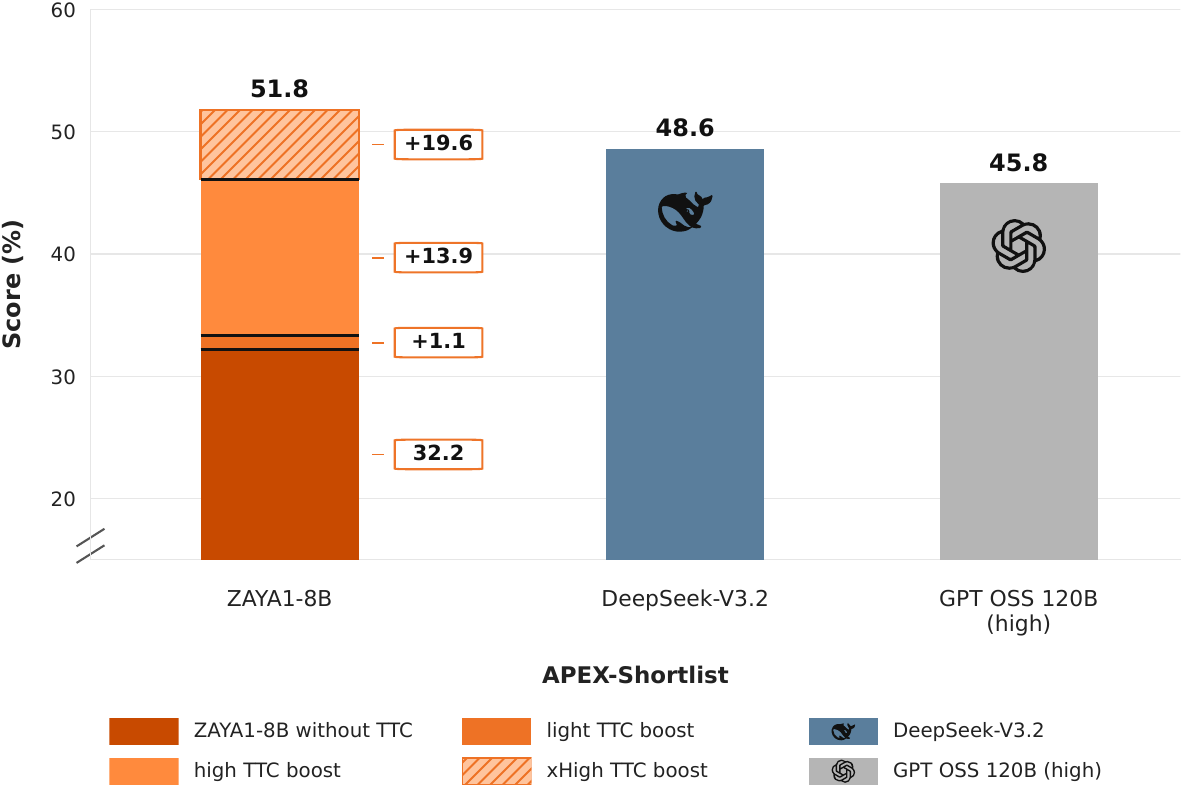}
    \caption{APEX-shortlist performance under the light, high, and extra-high Markovian RSA configurations from Table~\ref{tab:markovian-rsa-capacity-ceiling}. The extra-high configuration reaches 51.8\% using approximately 5.5M newly generated decode tokens per problem. GPT-OSS-120B (high) and DeepSeek-V3.2 comparator scores are taken from the MathArena leaderboard~\citep{dekoninck2026matharena}. These external comparator scores are shown for context only and may use different inference protocols.}
    \label{fig:markovian-rsa-apex}
\end{figure}

This should not be read as a general claim that tail length never matters. On harder benchmarks, the aggregation depth ($T$), the diversity of the candidate responses ($N$ and $C$), $\beta$, and $\tau$ can all contribute to saturating the model's capacity. We also evaluate higher-compute Markovian RSA settings on APEX-shortlist, a harder capacity-ceiling benchmark. Table~\ref{tab:markovian-rsa-capacity-ceiling} reports the three APEX configurations used for the light, high, and extra-high modes shown in Figure~\ref{fig:markovian-rsa-apex}. The strongest setting, with $T=8$, $N=32$, $C=4$, $\beta=32\mathrm{K}$, and $\tau=4\mathrm{K}$, reaches 51.8\% on APEX-shortlist. This setting uses approximately 5.5M newly generated decode tokens per problem, so we treat it as a capability-ceiling evaluation rather than a recommended deployment setting.

\begin{table}[h] 
    \centering 
    \small 
    \begin{tabular}{lccccc} 
    \toprule 
    $T$ & $N$ & $C$ & $\beta$ (CoT) & $\tau$ (Tail) & APEX-shortlist \\ 
    \midrule 
    2 & 16 & 4 & 8K & 4K & 33.3 \\ 
    \midrule 
    8 & 16 & 4 & 16K & 4K & 46.1\\ 
    8 & 32 & 4 & 32K & 4K & 51.8 \\ 
    \bottomrule 
    \end{tabular} 
    \caption{APEX-shortlist Markovian RSA configurations. These three rows define the light, high, and extra-high modes shown in Figure~\ref{fig:markovian-rsa-apex}.}
    \label{tab:markovian-rsa-capacity-ceiling} 
\end{table}

\paragraph{Markovian baseline and the value of aggregation} The $C=1$ row of Table~\ref{tab:markovian-rsa-scaling} runs the algorithm with a single previous tail conditioning each new candidate: each of the $N=16$ candidates carries forward only its own bounded textual state, with no cross-candidate aggregation. This is a parallel version of chunked Markovian-Thinker rollouts. Two observations follow. First, the $C=1$ baseline reaches 82.1\% on AIME'25 and 75.0\% on HMMT'25, indicating that bounded carryover preserves much of the model's long-CoT reasoning capability without aggregation --- consistent with \citep{markovian-thinker}'s finding that off-the-shelf reasoning models support Markovian traces zero-shot. Second, the gap between the $C=1$ baseline and the corresponding Markovian RSA configuration (4.4 points on AIME'25 and 5.8 points on HMMT'25, both at $\beta=8\text{K}$, $\tau=4\text{K}$) quantifies the additional gain from cross-candidate aggregation on top of bounded continuation. The two effects compose: bounded carryover establishes that the inference workload can be capped without losing reasoning capability, and recursive aggregation provides further gains from cross-candidate refinement.

\paragraph{Comparison to full-chain RSA}
Markovian RSA recovers full-chain RSA in the limit $\tau=\beta$, but the bounded-tail setting substantially reduces aggregation-prompt length. At $\beta=40\mathrm{K}$ and $C=4$, full-chain RSA would carry up to $160\mathrm{K}$ candidate-state tokens per aggregation item, whereas Markovian RSA with $\tau=4\mathrm{K}$ carries only $16\mathrm{K}$. We therefore use Markovian RSA for the reported high-budget evaluations. A fully matched empirical comparison against full-chain RSA is left for future work.

\begin{figure}[h]
    \centering
    \includegraphics[width=0.9\linewidth]{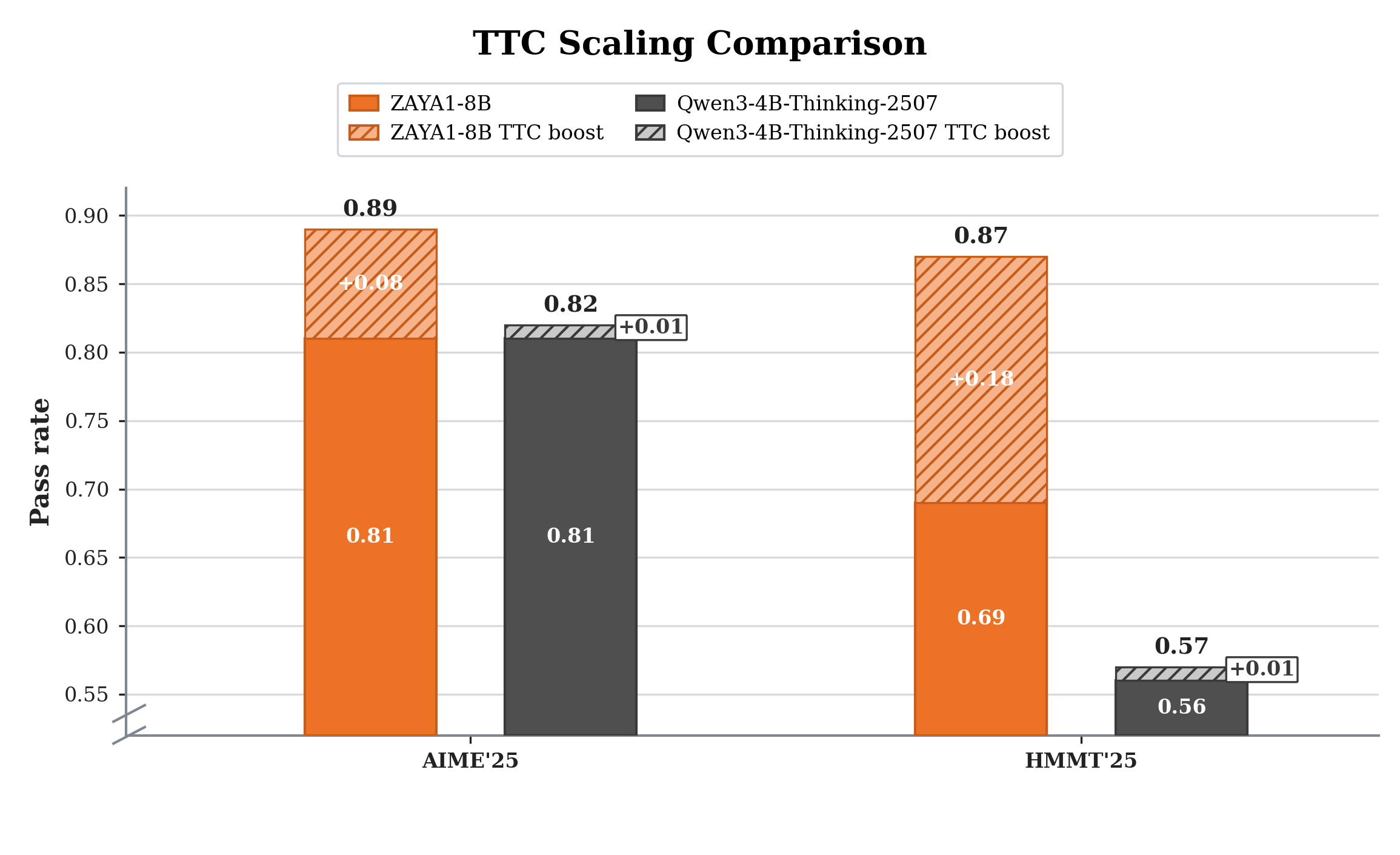}
    \caption{Test-time compute scaling for ZAYA1-8B and Qwen3-4B-Thinking-2507. Both models use the same TTC procedure (Markovian RSA with $T=2$, $\beta= 16\mathrm{K}$, and $\tau=4\mathrm{K}$) with the final response budget of 40K for both the TTC and non-TTC runs (the recommended sampling parameters in the model card were used for the Qwen model), so performance differences reflect model-level differences in the ability to exploit aggregation. ZAYA1-8B saw long-CoT reasoning data during midtraining and aggregation-based data during SFT and RL; Qwen3-4B-Thinking-2507 was not trained for this specific aggregation workflow. We interpret the gap as evidence of training-design effects rather than method differences.}
    \label{fig:ttc-cross-model}
\end{figure}

\paragraph{Cross-model comparison} Training for TTC. The benefits of training for a TTC workflow are not specific to Markovian RSA. Figure~\ref{fig:ttc-cross-model} compares ZAYA1-8B against Qwen3-4B-Thinking-2507 under Markovian RSA, with both models running the same TTC procedure. This comparison is not intended as a method evaluation, since both models use the same procedure. We interpret any gap as evidence of training-design effects rather than method differences: ZAYA1-8B is trained from midtraining onward with TTC aggregation traces in midtraining, SFT, and RL; Qwen3-4B-Thinking-2507 is not.

\paragraph{Serving profile and compute efficiency}
Markovian RSA has a favorable serving profile because every stage is both batched and bounded. Round-0 generation runs $N$ independent candidates in parallel. Later aggregation rounds also run at batch size $N$, with each item prefilling only the original problem plus $C$ carried-forward tails. With the deployment setting $N=16$, $C=4$, and $\tau=4$K, the candidate-state portion of each aggregation prompt is bounded by $C\tau=16$K tokens, plus the original problem and formatting overhead. Prefill is therefore bounded and predictable at every round, while decode proceeds as batched generation with per-candidate length capped by $\beta$.

This differs from both single long-CoT and full-chain RSA. A single long rollout has batch size $1$, with decode position and KV-cache length growing continuously with the full trace. Full-chain RSA batches candidates, but forwards full reasoning chains into aggregation prompts, so the carried candidate state grows with $C\beta$. At $\beta=40$K and $C=4$, full-chain RSA would carry up to $160$K candidate-state tokens per aggregation item, whereas Markovian RSA with $\tau=4$K carries only $16$K, a $90\%$ reduction in carried candidate state before accounting for the shared problem prompt. Markovian RSA therefore lets us increase per-candidate reasoning depth through $\beta$ without increasing aggregation-prefill length.

This bounded-context profile should be distinguished from total generated-token cost. Markovian RSA still spends a large aggregate decode budget because it generates many candidates across parallel workers and aggregation rounds. For the 40K/4K configuration, the realized total is approximately 740K newly generated decode tokens per problem across all workers on the reported AIME'25/HMMT'25 evaluation runs. We therefore compare TTC configurations using both active-parameter $\times$ total generated-token cost and the serving-side context profile: per-candidate decode capped by $\beta$, aggregation prefill bounded by $|q|+C\tau$, and no stage attending over the full generated reasoning history.

\paragraph{Recommended deployment configuration}
For deployment, we recommend the 16K/4K configuration from Table~\ref{tab:markovian-rsa-scaling} as a lower-cost default. It provides a strong accuracy--cost tradeoff while using substantially fewer realized decode tokens than 40K/4K: approximately 440K vs. 740K generated tokens per problem on the reported AIME'25/HMMT'25 evaluations, excluding prompt, prefill, and copied-tail tokens. In our current serving setup, we also observed the lightweight Markovian RSA configuration ($T=2, N=8, C=4, \beta=8\mathrm{K}, \tau=4\mathrm{K}$, and final response budget 40K) completing in roughly $0.4\times$ the wall-clock time of our standard $N=1$ long-reasoning baseline on the same evaluation harness. We attribute this to serving the workload as bounded-context batched decoding rather than as one long position-growing trace. We report this wall-clock ratio as an implementation-specific observation, not as a hardware-independent throughput benchmark. The 40K/4K configuration is reserved for capability-ceiling evaluation.

\section{Discussion}
\label{sec:discussion}

In this technical report, we presented ZAYA1-8B, the first and smallest model in the ZAYA-1 family. ZAYA1-8B is designed to maximize reasoning performance per active parameter, with a particular focus on reasoning-intensive mathematics and coding. In this target regime, the model is strongly competitive with systems that use far more active parameters, and reaches or exceeds the level of earlier frontier-scale reasoning models such as DeepSeek-R1-0528 and Gemini-2.5 Pro on several challenging math and code benchmarks.

With its native Markovian RSA TTC mode, ZAYA1-8B approaches the mathematical performance of much larger frontier reasoning models such as Gemini-2.5 Pro, DeepSeek-V3.2, and GPT-5-High. ZAYA1-8B is also competitive with substantially larger open-weight models including OLMo-3.1-32B-Think, Nemotron-3-Nano-30B-A3B, Intellect-3, and Mistral-Small-4-119B, with its clearest advantage on math/code reasoning density.

We attribute ZAYA1-8B's performance to a combination of its architecture, cascaded RL pipeline, reasoning-heavy training data, and an AMD training stack that supported long-context pretraining, midtraining, and SFT.

Finally, we introduced Markovian RSA, a test-time scaling method that combines recursive self-aggregation with bounded carry-forward state. The model generates and aggregates batches of candidate responses in parallel, while each aggregation round conditions only on a fixed number of bounded-length reasoning tails. This preserves RSA's cross-candidate refinement benefits while keeping aggregation context bounded, avoiding attention over the full generated reasoning history.

We believe TTC is an especially promising avenue for smaller reasoning-focused models and can make them competitive with substantially larger models in active-parameter $\times$ generated-token cost for some reasoning workloads. If TTC methods reliably convert additional generated tokens into accuracy, then active-parameter count and inference-time reasoning tokens become complementary axes of scaling.

Below, we discuss observations and lessons learned from the ZAYA1-8B training process.

\subsection{RL Sample Efficiency}
\label{sec:discussion-rl-efficiency}

The reasoning RL portion of ZAYA1-8B post-training is short relative to the pretraining and midtraining compute that precedes it. The main verifiable-reasoning cascade uses 232 reasoning-warmup steps, 400 RLVE-Gym steps, 384 math+code+TTC phase-1 steps, and 464 math+code+TTC phase-2 steps, for 1{,}480 total reasoning-RL update steps before behavioral RL. Despite this small number of optimizer steps, the aggregate post-training gain is large. Relative to the SFT checkpoint, we observe roughly a 20--30 point gain on AIME-like math evaluations and roughly a 10-point gain on LiveCodeBench-v6, with smaller but positive changes on many other benchmarks. Achieving gains of this size with ordinary midtraining or SFT would likely require substantially more data and optimization compute, and identifying the right supervised distribution would itself be nontrivial.

One interpretation is that RL takes a KL-minimal path to optimality during training \citep{shenfeld2025rlsrazor}, keeping the parameters relatively close to their initial values, compared to SFT, which may move the model arbitrarily in parameter space. Thus, the pretraining and midtraining stages appear to install most of the latent capabilities needed for reasoning; RL then changes the policy's sampling distribution so that these capabilities are expressed more reliably under long generation budgets. This view is consistent with our pass@k observations: in our previous base-model report, the reasoning checkpoint showed strong pass@64 behavior at a 30K generation budget, while the post-RL model moves much of that capability into average sampled performance at longer generation lengths. However, unlike prior work \citep{cui2025entropy} that argued RL simply uses up the inherent entropy of the midtraining checkpoint, we observed pass@k staying stable or even increasing during RL training where it did not hit the performance ceiling. 

The sample efficiency of RL remains an open problem. One plausible contributor is that the optimization problem in verifiable RL is qualitatively different from dense cross-entropy training. In SFT, every target token supplies a supervised gradient. In verifiable RL, useful signal is concentrated in trajectory-level outcomes, group-relative comparisons, verifier acceptance, and trust-region filtering: only a subset of prompts and sampled rollouts produces informative contrast for a given update. This makes the effective learning signal sparse over prompts and trajectories, before considering how the optimizer transforms that signal into parameter updates. We discuss optimizer-dependent sparsity in the resulting parameter deltas in Section~\ref{sec:discussion-momentum}.

We do not claim here that SFT on the same data would fail to recover the RL gains; we have not run this experiment in a controlled form for this release. The safer interpretation is that short verifiable-RL runs can move a strong pretrained/SFT model into a nearby region of policy space that is difficult to identify from supervised next-token prediction alone.

\subsection{Momentum-Free RL Optimization}
\label{sec:discussion-momentum}

ZAYA1-8B uses Muon with momentum set to zero for matrix-valued actor weights during RL. This is unconventional from the perspective of pretraining, where momentum and adaptive optimizer state are standard tools for stabilizing dense next-token-prediction training. In our RL setting, however, momentum-free Muon worked well and was compatible with the stability requirements of PipelineRL. We did not run a controlled optimizer ablation for this report, so we present this as an empirical recipe choice rather than as a general optimizer recommendation.

Our motivation is that asynchronous verifiable RL differs from pretraining in both signal structure and stationarity. Each actor update is tied to a rollout batch with sampled trajectories, verifier outcomes, group-relative advantages, and a generating policy snapshot that may differ from neighboring batches. In this setting, carrying first-moment state across batches can average gradient information collected under different prompt sets, sampled solutions, reward outcomes, and policy lags. This may be useful when the persistent component of the gradient dominates minibatch noise, but it may be less useful when the stationarity horizon of the RL signal is short relative to the optimizer's momentum horizon. We therefore use momentum-free Muon as a simple optimizer-state-reset variant: each actor update depends only on the current rollout batch while retaining Muon's normalized matrix update.

This choice is related to recent observations that RL finetuning of LLMs can update a relatively small subset of parameters and that memory-light optimizers can remain competitive in RL settings \citep{mukherjee2025rlsubnetworks,mukherjee2026sgdrl}. In an informal matched-step RL diagnostic, SGD updates left 99.51\% of parameters exactly unchanged and 99.94\% below $10^{-5}$, while AdamW left 92.82\% exactly unchanged and 96.40\% below $10^{-5}$. We do not use this diagnostic to claim that SGD, momentum-free Muon, or sparse updates are generally preferable, nor to claim that RL is universally sparse relative to SFT. Rather, it suggests that in our setting the effective update can be highly concentrated, making optimizer-state design a relevant practical consideration.

Momentum-free Muon also reduces optimizer-state memory because no persistent Muon first-moment buffer is maintained for the actor weights. For matrix-valued actor parameters, the update remains Muon's normalized matrix update rather than a raw SGD step, while embedding and output-head parameters are optimized with AdamW under the standard matrix-parameter split. A direct comparison against momentum Muon, AdamW, and SGD actor updates is left for future work.

\subsection{MoE Logit Mismatch and Router Replay}
\label{sec:discussion-router-replay}

PipelineRL requires the gradient step to be computed against the same policy distribution that generated the rollout, up to bounded staleness. In practice, this is an SNR problem rather than a binary correctness problem. Small engine-trainer logit differences add noise to the policy-gradient estimate; when the rest of the recipe is stable, this noise tends to appear as slower learning, unstable high-learning-rate behavior, or plateaus rather than immediate collapse.

MoE models introduce an additional source of mismatch beyond ordinary numerical differences between the rollout engine and trainer. In a dense model, a small numeric perturbation usually causes a small logit perturbation. In a top-1 MoE, the same perturbation can flip a token's expert assignment, producing a discontinuous change in the computation path. A token generated by expert $e_{\mathrm{rollout}}$ but trained through expert $e_{\mathrm{train}} \neq e_{\mathrm{rollout}}$ gives the actor gradient the wrong local model for that token. This mismatch is especially harmful in long-rollout RL, where many such token-level routing errors can accumulate across a sequence.

Router replay \citep{ma2025stabilizingmoereinforcementlearning} addresses this MoE-specific mismatch by recording the per-token, per-layer expert choices made by vLLM during rollout generation and replaying those choices during all trainer forward passes over the rollout. In ZAYA1-8B training, router replay was a major stability improvement: high learning rates that were unstable without replay became usable with replay. Router replay does not remove all sources of mismatch, but it removes the discontinuous top-1 routing component.

The remaining engine-trainer mismatch is handled by the FP32 operation set described in Section~\ref{sec:precision}. We use engine-vs-trainer probability scatter plots as the main diagnostic: before hardening and router replay, the scatter broadens and token probabilities deviate from the identity line; after hardening, the distributions align closely. In the final configuration, the engine-trainer comparison reaches KL divergence approximately $1.3 \times 10^{-4}$ and Pearson correlation above $0.9996$ on a 128-prompt, $G=16$, 4K-completion diagnostic batch. We found substantial benefits in terms of training stability and final performance from being extremely careful about reducing sources of numerical error, even if they initially seemed small.

\subsection{Data and Verifier SNR}
\label{sec:discussion-snr}

The sample efficiency of verifiable RL depends strongly on data and verifier SNR. A verifier can provide binary reward at scale, but a binary reward is only useful when the prompt distribution produces informative variation across sampled rollouts. If most groups are solved by every rollout, the gradient has little contrast. If no rollout solves the prompt, the gradient is also weak. If the verifier accepts shortcuts or rewards a skewed answer distribution, the model can learn the shortcut rather than the intended reasoning behavior which is often described as `reward hacking'.

We therefore treat difficulty curation as a fundamental part of the RL algorithm rather than as a preprocessing detail. The reasoning warmup and math+code+TTC stages use pass-rate filtering to select hard but not fully saturated prompts. The RLVE-Gym stage performs this filtering online through an adaptive difficulty scheduler targeting the high-information region of each environment. Between major stages, we re-filter our datasets using the current RL policy: first with an instruction-mode filtering pass using more aggressive sampling settings, and then with a thinking-mode pass closer to the RL rollout setup but at a lower response-length limit. These filtering passes remove the high end of the distribution, progressively raising the difficulty floor as the model improves.

Verifier quality becomes more important as the model approaches the ceiling of a benchmark or training distribution. At low capability, even a coarse verifier can provide useful signal because the model is far from saturation. Near saturation, small verifier errors, spurious binary ground-truth patterns, incomplete code tests, or skewed answer distributions can dominate the remaining gradient signal, often causing learning to plateau. In practice, we typically could trace back a plateau to one of three causes: prompts that were too easy, prompts that were effectively impossible under the current rollout budget, or verifier/data artifacts that made a shortcut easier than genuine reasoning.

This observation also helps explain why short RL runs can be effective. When the prompt distribution is centered near the model's current capability boundary and the verifier is high precision, each rollout batch contains useful contrast. When data or verifier SNR degrades, increasing RL steps alone is inefficient: the optimizer repeatedly sees low-information or misleading groups which ultimately diminish the signal to the point where learning plateaus at a noise floor. For this reason, the ZAYA1-8B cascade alternates capability-building stages with data curation and difficulty filtering, rather than treating the RL dataset as fixed.

\subsection{Why ZAYA1-8B Benefits from Test-Time Compute}
\label{sec:discussion-smoe-ttc}

ZAYA1-8B is a small-active-parameter MoE: each generated token uses roughly 700M active parameters while drawing on a larger 8B-parameter expert pool across tokens. This makes test-time compute especially attractive. Generating many candidate traces is cheap in active-parameter compute, while the total expert pool still gives the model more specialization capacity than a similarly active dense model. Inference compute is therefore better measured by active parameters multiplied by generated tokens than by total parameters alone.

Markovian RSA exploits this regime. The method spends additional compute on many relatively cheap candidate rollouts and aggregation passes, while bounding aggregation context through the carried-forward tail. This lets ZAYA1-8B trade more decode tokens for higher accuracy without paying the per-token cost of a much larger dense or high-active-parameter model. The relevant deployment question is not just whether accuracy increases with more tokens, but whether the active-parameter $\times$ generated-token product is favorable relative to larger alternatives.

Training also matters. ZAYA1-8B sees TTC aggregation traces before inference: long-CoT data appears during midtraining, aggregation-based examples appear during SFT, and Markovian RSA prompts are included in the math+code+TTC RL stage. As a result, the model is not asked to discover aggregation behavior only at inference time. It has learned a strong prior for reading several candidate tails, reconciling partial reasoning paths, and producing an improved solution.

This combination helps explain why Markovian RSA scales well on ZAYA1-8B. The model has low per-token active compute, enough total expert capacity to support diverse candidate trajectories, and explicit training exposure to the aggregation workflow. We have observed weaker TTC scaling from models not trained for this workflow under the same RSA, Markovian RSA, and PaCoRe procedures, but those comparisons are not fully controlled because the models differ in architecture, training data, and post-training recipe. We therefore treat the main claim as specific to ZAYA1-8B: its architecture and training recipe make test-time compute a particularly effective way to convert low-active-parameter rollouts into higher reasoning accuracy.

\subsection{KL-in-Reward and Length Bias under PipelineRL}
\label{sec:discussion-kl-length}

Section~\ref{sec:stability} describes a length-bias failure mode we observed when combining PipelineRL with a sampled signed $K_1$-estimator log-ratio term in the reward. In short, stale or mixed-policy rollouts can make the signed sequence-level log-ratio negative, so subtracting it from reward can create a positive length-dependent offset. When one completion spans multiple generator-policy snapshots, stale-prefix terms can also affect the sequence-level advantage assigned to fresher suffix tokens. For ZAYA1-8B, we avoided this configuration by removing KL-in-reward and relying on DPPO Binary-TV for trust-region control. More principled chunk-local signed-log-ratio handling and staleness rescaling are left for future work.

\subsection{Open Questions and Limitations}
\label{sec:discussion-limitations}

Several open questions remain. We provide evidence for the viability of the AMD hardware and networking stack for pretraining at the 8B scale, which is larger than prior public pretraining runs on combined AMD GPU and networking hardware that we are aware of. However, training ZAYA1-8B required only data parallelism plus context parallelism during context-extension phases. Although we have stress-tested other parallelism strategies, including cross-node parallelism, further scaling is needed to validate the stack for substantially larger models.

It was also unclear at the outset whether our architectural changes would support effective reasoning and long-context behavior. This was a particular concern for CCA, which we had not previously tested at long contexts. ZAYA1-8B's performance on long-context and reasoning benchmarks suggests that CCA's advantage over attention variants such as MLA and GQA can be maintained in these settings. However, 8B total parameters is still modest relative to frontier-scale models, and architectural behavior at larger scales remains to be tested.

The evaluation profile of ZAYA1-8B is uneven in a useful way. The model is strongest on reasoning-heavy mathematics and code, where it is competitive with much larger models and, under TTC, approaches frontier mathematical performance. On knowledge-heavy and broad factual evaluations such as MMLU-Pro and GPQA-D, ZAYA1-8B remains strong for its active-parameter scale but does not fully close the gap to substantially larger models. This pattern is consistent with the intuition that reasoning performance and factual storage scale differently: a small-active model can express strong algorithmic reasoning while still having less capacity for broad memorized knowledge than much larger models.

This motivates a useful direction for future systems: small-active reasoning models may be especially effective when paired with test-time compute and external retrieval. Rather than requiring all capability to be stored in parameters, such systems can combine a compact reasoning core, cheap parallel inference, and external knowledge sources. ZAYA1-8B is one example of this tradeoff, but establishing the generality of this pattern requires further scaling and controlled comparisons.

Some of ZAYA1-8B's reasoning strength may come from its relatively deep architecture. ZAYA1-8B has 40 layers, compared with 36 layers for Qwen3-4B and 16 layers for OLMoE at a similar total-parameter scale. We hypothesize that this depth helps the model represent more serial computation within a single forward pass, which may be useful for reasoning. The residual-scaling mechanism may also help preserve the contribution of later layers by controlling residual-norm growth, while the ZAYA1 router improves routing stability and expert specialization. We treat these as architectural hypotheses supported by our training experience rather than as isolated causal claims; controlled ablations at larger scale are left for future work.

On agentic tasks, ZAYA1-8B trails models whose post-training emphasizes multi-turn tool use, especially on benchmarks such as BFCL-v4 and $\tau^2$. This is an expected consequence of the release scope: our midtraining, SFT, and RL budgets prioritize math, code, TTC aggregation, and instruction following rather than dedicated multi-turn agentic RL. We view this as a data-and-training emphasis gap rather than an architectural limitation. Scaling agentic data and RL remains a priority for future releases.



\section*{Acknowledgements}

We would like to thank Paul White, Danny Martinelli, Steven Brook, Kristina Zhao, and Krithik Puthalath for help with the release. We would also like to thank Yuankai Chen and Yao Fu from AMD for their support and close technical collaboration.


\bibliographystyle{tmlr}
\bibliography{main}

\clearpage
\appendices

\section{Cluster Details}
\label{app:cluster}

Table~\ref{tab:hardware-config} summarizes the hardware configuration of the
compute, storage, and login nodes. All nodes also include separate local drives
for the operating system.

\begin{table*}[t]
\centering
\small
\setlength{\tabcolsep}{5pt}
\begin{tabularx}{\textwidth}{@{}l l X X@{}}
\toprule
\textbf{Node} & \textbf{Component} & \textbf{Specification} & \textbf{Details} \\
\midrule

\multirow{5}{*}{Compute}
& GPUs
& $8\times$ AMD MI300X GPUs~\citep{amd2024cdna3}
& Connected via Infinity Fabric intra-node interconnect. \\

& RAM
& 2\,TB DDR5
& $16\times128$\,GB Samsung M321RAJA0MB0-CWMNY DIMMs running at 5600\,MT/s. \\

& CPU
& Dual-socket Intel Xeon Platinum 8570
& Each socket has 56 physical cores and 2 threads per core, and is connected to 1\,TB of RAM ($8$ DIMMs). \\

& Networking
& $8\times$ Pollara 400 NICs~\citep{amd_pollara_2024} + 1 Pensando DSC 200\,GbE NIC
& Each Pollara NIC provides 400\,Gb/s; the Pensando NIC is used for data and checkpoint transfer. \\

& Storage
& 25.6\,TB NVMe
& $8\times$ Micron MTFDKCC3T2TGQ-1BK1DABDB drives, 3.2\,TB each. \\

\midrule

\multirow{4}{*}{Storage}
& RAM
& 256\,GB DDR5
& $16\times16$\,GB Samsung M321R2GA3BB6-CQKET DIMMs running at 4800\,MT/s. \\

& CPU
& Dual-socket Intel Xeon Gold 6426Y
& Each socket has 16 physical cores and 2 threads per core, and is connected to 128\,GB of RAM ($8$ DIMMs). \\

& Networking
& 1 Pensando DSC 100\,GbE NIC
& Used for data and checkpoint transfer. \\

& Storage
& 120\,TB RAID0
& Single RAID0 array, \texttt{/dev/md127}, built from $16\times$ Micron 7450 MTFDKCC7T6TFR NVMe drives, each with approximately 7.6\,TB. \\

\midrule

\multirow{3}{*}{Login}
& RAM
& 80\,GB system memory
& $5\times16$\,GB DIMMs under QEMU. \\

& CPU
& Virtualized dual-socket Intel Xeon (Sapphire Rapids)
& Each socket has 8 cores and 2 threads per core; virtualized under KVM. \\

& Storage
& Approximately 1\,TB
& Three virtual disks: \texttt{vda} (100\,GB), \texttt{vdb} (520\,GB), and \texttt{vdc} (520\,GB). \\

\bottomrule
\end{tabularx}
\caption{Hardware configuration of the compute, storage, and login nodes.}
\label{tab:hardware-config}
\end{table*}

\section{Storage Node Sizing and I/O Calculations}
\label{app:storage-io}

We analyze shared-storage requirements for dataset reads during training, assuming Megatron-style pretokenized corpora accessed through \texttt{mmap} or buffered reads on a dedicated storage fabric. Large sequential checkpoint writes are throughput-bound rather than IOPS-bound and are not considered here.

Let $G$ be the global batch size, $s$ the sequence length, $b$ bytes per token, $P$ the storage page size, $t$ the iteration time, and $I_{\max}$ the sustainable IOPS capacity. We introduce a \emph{scatter factor} $\sigma \ge 1$ to capture how much real dataset access patterns, including metadata touches, index probes, page-cache misses, and small random seeks, deviate from perfectly contiguous reads. Each iteration reads $G \cdot s \cdot b$ bytes, requiring
\begin{align}
N_{\text{IO/iter}}
\;=\;
\sigma \cdot \left\lceil \frac{G \cdot s \cdot b}{P} \right\rceil
\end{align}
effective I/O operations. The sustained IOPS requirement is therefore
\begin{align}
\text{IOPS}_{\text{needed}}
\;=\;\frac{N_{\text{IO/iter}}}{t}\;=\;
\frac{\sigma}{t}\cdot \left\lceil \frac{G \cdot s \cdot b}{P} \right\rceil ,
\label{eq:iops-needed}
\end{align}
and the break-even iteration time under budget $I_{\max}$ is
\begin{align}
t_{\text{break}}
\;=\;
\frac{\sigma}{I_{\max}}\cdot \left\lceil \frac{G \cdot s \cdot b}{P} \right\rceil .
\label{eq:t-break}
\end{align}

We can estimate the scatter factor $\sigma$ from the average number of additional page faults per sample. Let $m$ denote the average count of additional page faults from metadata, \texttt{*.idx} probes, document-boundary straddles, or cold reads. If the ideal pages per sample are $(s \cdot b)/P$, a practical approximation is
\begin{align}
\sigma
\;\approx\;
1 \;+\; \frac{m \cdot P}{s \cdot b}.
\label{eq:sigma}
\end{align}
This interpolates between contiguous, warm-cache access ($\sigma \to 1$) and fragmented, small-document regimes ($\sigma > 1$). In our experience with well-packed Megatron datasets, $\sigma \in [1,2]$ is typical; heavily fragmented or multi-shard random-seek workloads can reach $\sigma \in [2,8]$.

For the ZAYA1 training run with $G{=}4096$, $s{=}4096$, $b{=}4$~B, $P{=}4096$~B, $t{=}2.5$~s, and $I_{\max}{=}70{,}000$ IOPS, each iteration reads 64~MiB across 16{,}384 pages. This requires approximately $6{,}554 \cdot \sigma$ IOPS, with break-even time $t_{\text{break}} \approx 0.234 \cdot \sigma$~s. At the observed $t{=}2.5$~s, the run remains above the break-even time even for $\sigma{=}8$, indicating that the 70K IOPS storage budget is sufficient.

\section{Compressed Convolutional Attention}
\label{app:cca}

\begin{figure}[htbp]
    \centering
    \includegraphics[width=0.45\textwidth]{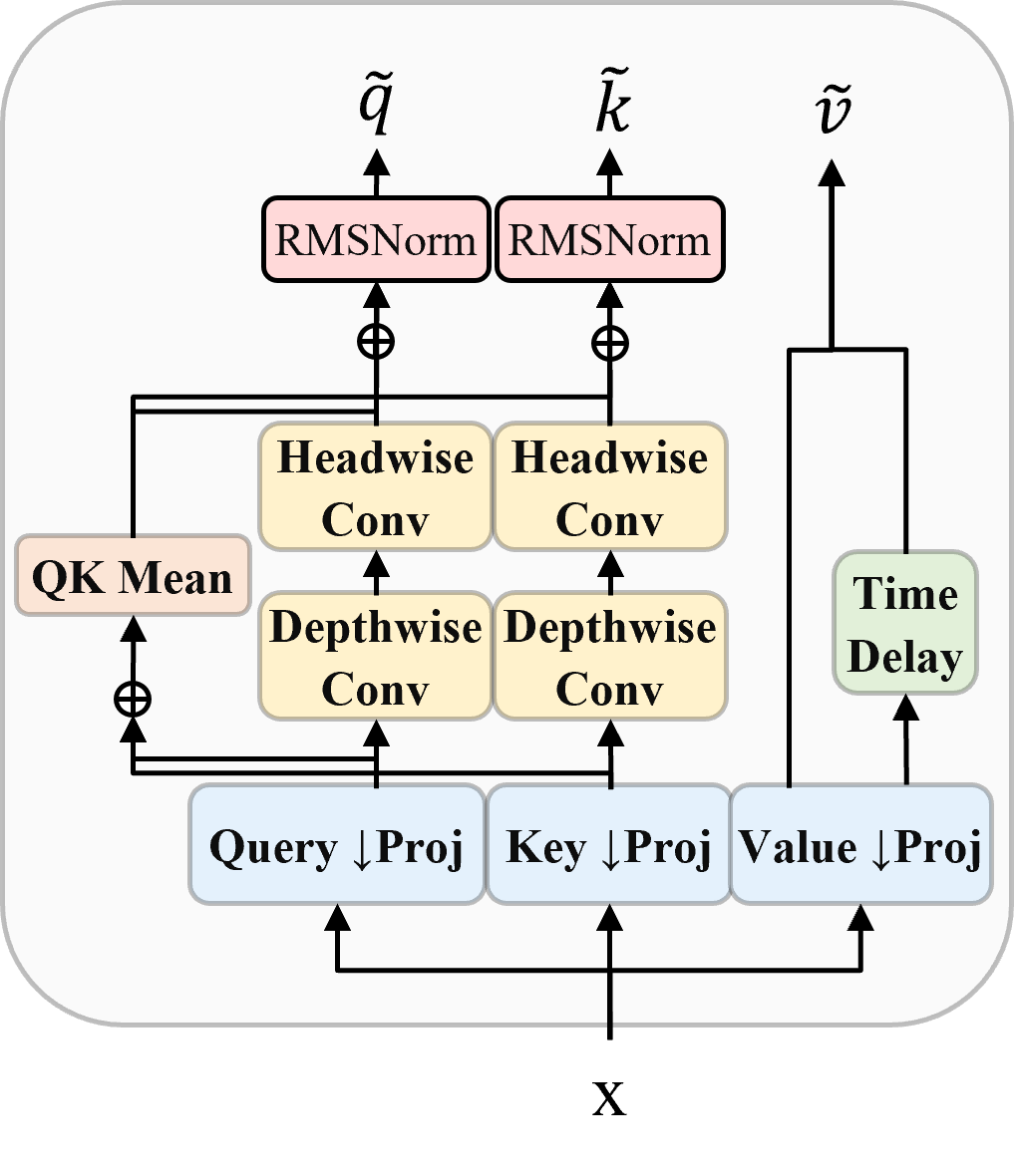}
    \caption{We briefly review the Compressed Convolutional Attention block \citep{cca}.}
    \label{fig:cca}
\end{figure}
CCA (shown in Figure~\ref{fig:cca}) modifies the attention block so that attention is performed in a compressed latent space. This reduces both memory and FLOP costs. In prior work, CCA outperformed alternatives such as GQA and MLA on perplexity and training/inference FLOPs while enabling high KV-cache compression, which is important for fast decoding.

CCA has several core components:
\begin{itemize}
    \setlength{\itemsep}{0pt}
    \item \textbf{Low-Rank Projections:} Low-rank down-projections reduce compute and memory.
    \item \textbf{Sequence-Mixing Convolutions:} A short convolution and grouped head-wise convolution act as lightweight preconditioners before attention.
    \item \textbf{Value Head Time-Delay:} A time delay of one token is applied to half of the value heads.
    \item \textbf{Skip Connections and Normalization:} The architecture utilizes query/key mean skip connections to enforce representational similarity, coupled with an RMSNorm layer that applies a head-wise temperature strictly to the keys. To stabilize training, we modify the standard CCA mechanism by scaling the keys by a learned temperature $T$ instead of $\exp(T)$. Because $\exp(T)$ can easily grow excessively large, the query-key inner product becomes susceptible to unbounded growth. Our linear parameterization successfully mitigates the resulting maximum attention logit instability, drawing parallels to similar stabilization efforts in MLA \citep{kimiteam2026kimik2openagentic}. The scaled query-key inner product is then upper bounded (assuming QK norm) to:
\begin{align}
    \frac{T \cdot d_h}{\sqrt{d_h}} = T \sqrt{d_h}\,,
\end{align}
where $d_h$ is the head dimension. 
\end{itemize}

These additions to precondition query and key, especially the convolutions, provide expressivity and nonlinearity that allow CCA to match or exceed full attention while requiring less compute and memory.

The CCA block can also operate in `GQA-mode', where multiple KV heads are shared across query heads. This further reduces decoding cost through additional KV-cache compression. We call the combined method CCGQA. For ZAYA1-8B, we use CCGQA with 2 KV heads for 8 query heads, on top of $2\times$ query compression, for an $8\times$ KV-cache compression ratio relative to full multi-head attention. 

\section{Expert Redundancy Diagnostic}
\label{app:expert-subspace-overlap}

We include a small diagnostic for expert redundancy in the MoE feed-forward blocks. This diagnostic is not an evaluation benchmark and should not be interpreted as a complete measure of expert specialization. Its purpose is narrower: to check whether ZAYA1-8B's experts appear unusually collapsed or redundant relative to other public MoE checkpoints.

For each MoE layer $l$ and expert $e$, let $W_{l,e}$ denote the expert projection under study. We compute the top-$d$ singular subspace $Q_{l,e} \in \mathbb{R}^{D \times d}$ with $d=128$ and orthonormal columns. For the input-side metric, $Q_{l,e}$ is the right singular subspace of the full first FFN projection. For gated FFNs with separate branches, we concatenate the gate and up projections along the FFN dimension before computing the right singular subspace; for ZAYA1-8B, this corresponds to the fused \texttt{linear\_fc1} projection. For the output-side metric, $Q_{l,e}$ is the left singular subspace of the expert output projection, corresponding to \texttt{linear\_fc2}, \texttt{down\_proj}, or \texttt{w2}.

For two experts $i$ and $j$ in the same layer, we define
\begin{equation}
    s_{l,i,j}
    =
    \frac{1}{d}
    \left\| Q_{l,i}^{\top} Q_{l,j} \right\|_{F}^{2}\,,
\end{equation}
where $\|\cdot\|_F$ is the Frobenius norm. The reported score averages $s_{l,i,j}$ over all off-diagonal expert pairs and all MoE layers. Larger values indicate that experts share more of the same input or output directions; smaller values indicate more distinct expert subspaces under this particular projection-space diagnostic.
\begin{table*}[ht!]
\centering
\small
\setlength{\tabcolsep}{3.5pt}
\begin{tabular}{lrrcccccc}
\toprule
Model & MoE layers & Experts/layer & Input overlap & Input $\rho$ & Input Var. & Output overlap & Output $\rho$ & Output Var. \\
\midrule
ZAYA1-8B & 40 & 16  & $0.0904 \pm 0.0244$ & $1.45\times$ & 17.8\% & $0.0990 \pm 0.0289$ & $1.58\times$ & 22.7\% \\
LFM2-8B-A1B & 22 & 32  & $0.1328 \pm 0.0502$ & $2.12\times$ & 25.4\% & $0.1174 \pm 0.0441$ & $1.88\times$ & 28.5\% \\
OLMoE-1B-7B & 16 & 64  & $0.1031 \pm 0.0397$ & $1.65\times$ & 29.7\% & $0.0872 \pm 0.0256$ & $1.40\times$ & 34.0\% \\
Qwen3-30B-A3B & 48 & 128 & $0.0923 \pm 0.0333$ & $1.48\times$ & 31.2\% & $0.0851 \pm 0.0226$ & $1.36\times$ & 39.2\% \\
\bottomrule
\end{tabular}
\caption{Mean within-layer expert subspace overlap with $d=128$. Input overlap uses the right singular subspace of the full first FFN projection; output overlap uses the left singular subspace of the expert output projection. Standard deviations are over all off-diagonal expert pairs across analyzed MoE layers. All rows use a 2048-dimensional comparison space, giving a random-subspace baseline of $d/D=0.0625$. The Qwen row uses Qwen3-30B-A3B-Thinking-2507.}
\label{tab:expert-subspace-overlap}
\end{table*}
\begin{figure*}[t]
    \centering
    \includegraphics[width=\linewidth]{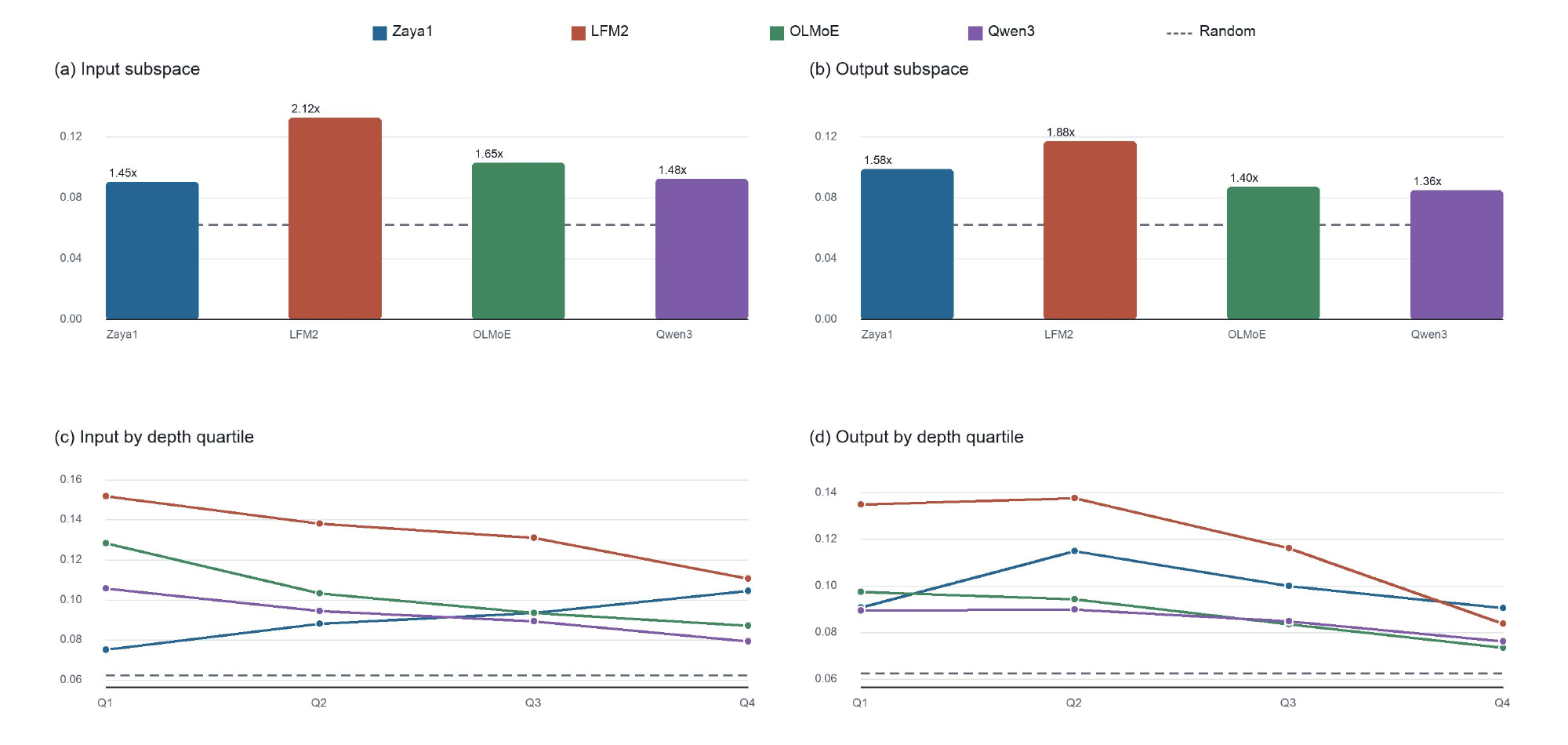}
    \caption{Expert redundancy diagnostic. Panels (a) and (b) show global mean overlap for input and output projections; bar labels report the random-normalized ratio $\rho$. Panels (c) and (d) show raw overlap averaged over depth quartiles. The dashed horizontal line is the random-subspace baseline $d/D=0.0625$. ZAYA1-8B is not an outlier toward higher expert overlap: its first-projection input overlap is close to Qwen and below LFM2 and OLMoE, while its output-projection overlap is intermediate between LFM2 and the lower-overlap OLMoE/Qwen group.}
    \label{fig:expert-subspace-overlap}
\end{figure*}
For independent random $d$-dimensional subspaces in $\mathbb{R}^{D}$, the expected value of this metric is $d/D$. All models in Table~\ref{tab:expert-subspace-overlap} have $D=2048$ for both the input and output comparison spaces, so the random-subspace floor is $128/2048 = 0.0625$. We therefore report both the raw overlap and the random-normalized ratio
\begin{equation}
    \rho = \frac{s}{d/D}.
\end{equation}
The ``Var.'' columns report the mean fraction of projection Frobenius energy captured by the top-128 singular directions; they are included as spectral context, not as a normalization of the overlap score.

Overall, this diagnostic does not indicate unusual expert collapse in ZAYA1-8B. On the input-side first projection, ZAYA1-8B is $1.45\times$ the random-subspace baseline, close to Qwen's $1.48\times$ and below LFM2 and OLMoE. On the output projection, ZAYA1-8B is intermediate at $1.58\times$ random: higher than OLMoE and Qwen, but below LFM2. ZAYA1-8B's input-side overlap also rises with depth, from $0.075$ in the first depth quartile to $0.105$ in the last, indicating more shared input directions in later MoE layers. We therefore use this appendix only as evidence against obvious expert collapse, not as a claim that ZAYA1-8B has stronger output-side specialization than all baselines.

\section{Rare-token sampling artifacts}
\label{app:gibberish}

We define gibberish as brief, recoverable episodes of seemingly random tokens that interrupt the CoT, and we argue that it is distinct from other forms of degeneration, which may involve cache corruption, numeric issues, repeats, etc. because these other issues are generally non-recoverable or at least persistent in the CoT.

In this set of examples (Table \ref{tab:gibberish_examples}), it is clear that the gibberish token has no relationship to the prefix, and interestingly, the suffix seems to mostly ignore it. However, this does not mean it is benign, because we found that it is likely to reinforce itself as a result of successful (rewarded) traces containing gibberish tokens.

\begin{table*}[h!]
\centering
\scriptsize
\setlength{\tabcolsep}{4pt}
\renewcommand{\arraystretch}{1.00}

\resizebox{\textwidth}{!}{%
\begin{tabular}{@{}r r c l@{}}
\toprule
\# & \textbf{Prefix} & \textbf{Gibberish} & \textbf{Suffix} \\
\midrule
0
& Let ' s count underscores : -- The outer parentheses : open before underscore : \{\} - Inside
& \texttt{<unused5148>}
& ? Actually start of right part : ( \_ | ( \_ \& ( \\

1
& , we have : - first underscore : c - second underscore : inside the '\& '? Actually
& \texttt{U+1490}
& . Wait after the " | (" we have '\_ -> maybe a variable \\

2
& as : ( c | ( d \& ( e \textasciicircum{} ( f | g )))) . Then outer parentheses
& \texttt{U+96BC}
& might not be needed . Thus we have outer parentheses around the \\

3
& (\_ | \_) )))) ` which we can view as '(' followed by '\_ then '| ' then '(' then
& \texttt{<unused521>}
& etc followed by ')) '? Actually need to count correctly . Let ' \\

4
& then '\textasciicircum{} ', then underscore , then ")" -> you have an outer parentheses containing
& \texttt{U+52DF}
& ?). That yields the expression "(" \_ \textasciicircum{} \_ ") "? That seems odd \\

5
& the outer parentheses of the XOR is the one after "(" , we have "(" underscore "\textasciicircum{} " then
& \texttt{U+12D6}
& ? Let ' s expression parse : The substring is " (\_ \textasciicircum{} (\_ \\

6
& just "(" "\_ ", "\textasciicircum{} ", "\_" ") ". That ' s parentheses open and close on a level containing
& \texttt{U+78C1}
& ? Possibly they just open parentheses for each underscore : '(' then \\
\bottomrule
\end{tabular}
}
\caption{Traces at the gibberish token position.}
\label{tab:gibberish_examples}
\end{table*}

As a heuristic for gibberish detection we rely on the method described in \citep{copet2025cwm} which combines a low logprob mask (2 nats below uniform) with token ID mask (a proxy for token rarity). While this is a good indicator/metric in practice, it is a somewhat arbitrary hard threshold and in our experience it isn't precise or sensitive enough to be used in training for reward zeroing or loss masking.

\begin{figure}[!t]
\centering
\includegraphics[width=\columnwidth]{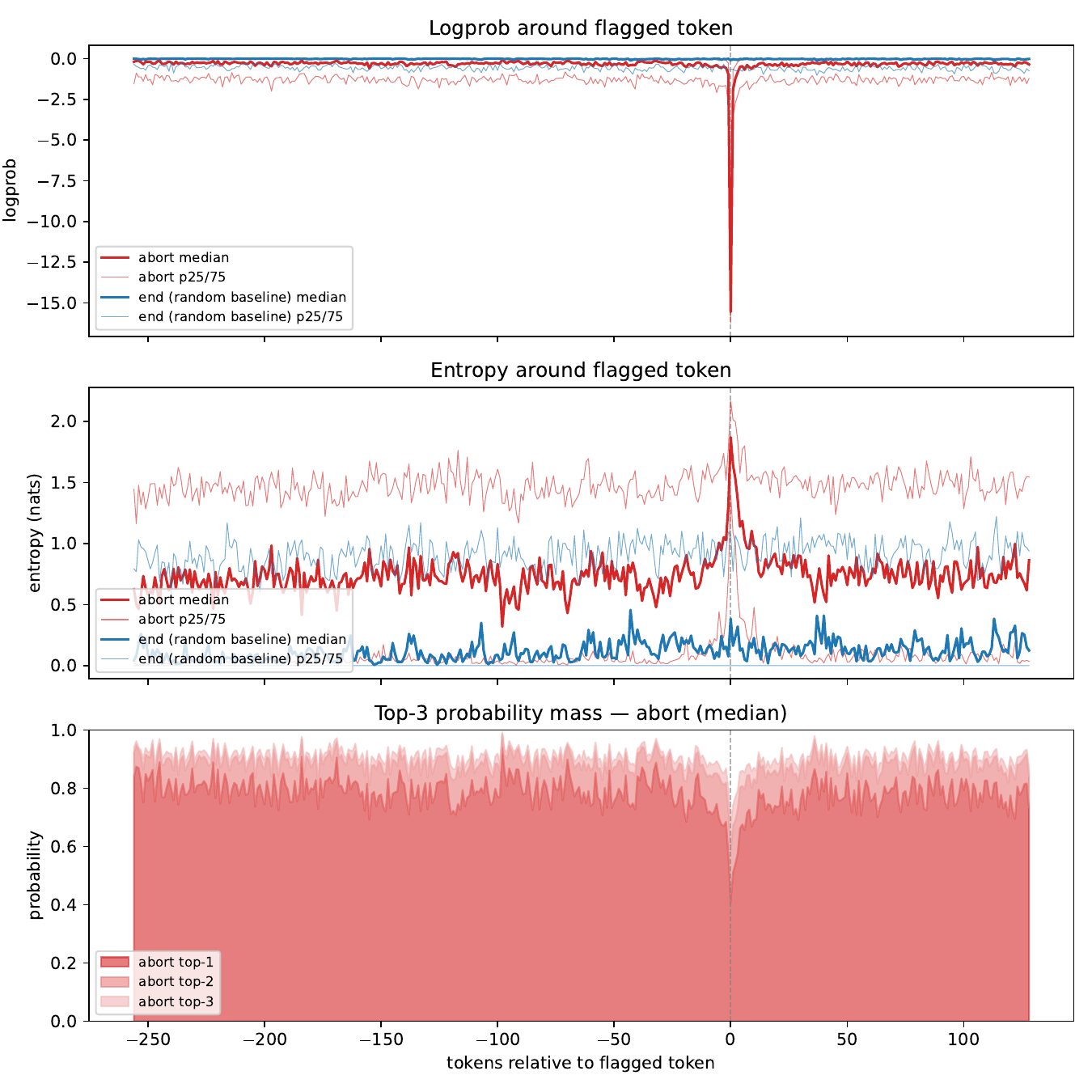}
\caption{Aggregated tokenwise metrics for gibberish responses compared to normal responses.}
\label{fig:gibberish-entropy}
\end{figure}

\begin{figure}[!t]
\centering
\includegraphics[width=\columnwidth]{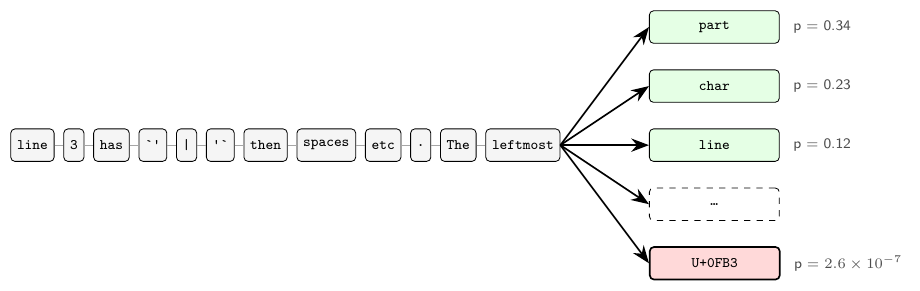}
\caption{Top token choices for an otherwise coherent trace that sampled a random Tibetan token. In this example the model was attempting to reason about a challenging isometric ASCII drawing problem from the RLVE gym known as BlockImage.}
\label{fig:gibberish-top-tokens}
\end{figure}

We collected traces and their top-3 token probabilities using the vLLM streaming API and aborted the traces around any gibberish-flagged tokens as they arose using the heuristic mask. While entropy is spiking in those cases, the most probable tokens still constitute a significant portion of the total probability. Looking into the top-$k$ tokens themselves, we found that the top token was a coherent continuation of the prefix, with the most common case being simply "the" as with examples 0, 1, 4, 6 in the table above. In another more context-dependent case, any of the top 3 most likely tokens would have been a coherent alternative (Figure \ref{fig:gibberish-top-tokens}).

In general, the baseline/prefix entropy of traces aborted with gibberish tokens tends to be higher (Figure \ref{fig:gibberish-entropy}). This could be because the model is challenged by a difficult problem and not confident as a result. Around the flagged token, entropy spikes, and while the probability mass in the top 3 tokens is median ~0.7, a significant mass remains in the tail of the distribution. For this reason, we believe that despite the sampled token having an extremely low probability, it was chosen because it was one of many possible low-probability tokens.

A natural solution is to use min-$p$ \citep{nguyen2024turning} sampling to eliminate extremely low probability choices from the sampling distribution. In our experiments, we find that even in the most degraded test checkpoint available to us, with a baseline rate of 19.9\% of flagged responses, min-$p$ sampling with a threshold of $10^{-5}$ reduces this to 0.1\%. In training, to prevent engine/trainer mismatch, we recommend implementing min-$p$ replay, which exposes the kept/omitted token IDs to the trainer for consistency in the min-$p$ renormalization.

\end{document}